\batchmode
\makeatletter
\def\input@path{{/media/data/important/doneWork/myWorks/isprs_2013//}}
\makeatother
\documentclass[11pt,english,onecolumn,twoside]{IEEEtran}
\usepackage[T1]{fontenc}
\usepackage[utf8]{inputenc}
\usepackage{geometry}
\usepackage{xcolor}
\usepackage{pdfcolmk}
\definecolor{page_backgroundcolor}{rgb}{1, 1, 1}
\pagecolor{page_backgroundcolor}
\usepackage{babel}
\usepackage{array}
\usepackage{float}
\usepackage{url}
\usepackage{multirow}
\usepackage{amsthm}
\usepackage{amstext}
\usepackage{amssymb}
\usepackage{graphicx}
\usepackage{setspace}
\PassOptionsToPackage{normalem}{ulem}
\usepackage{ulem}
\usepackage[unicode=true,
 bookmarks=true,bookmarksnumbered=false,bookmarksopen=false,
 breaklinks=false,pdfborder={0 0 0},backref=false,colorlinks=true]
 {hyperref}
\hypersetup{pdftitle={Structured Sparse Method for Hyperspectral Unmixing},
 pdfauthor={Fei-yun Zhu},
 citecolor=blue, urlcolor=magenta}

\makeatletter

\providecommand{\tabularnewline}{\\}
\floatstyle{ruled}
\newfloat{algorithm}{tbp}{loa}
\providecommand{\algorithmname}{Algorithm}
\floatname{algorithm}{\protect\algorithmname}
\providecolor{lyxadded}{rgb}{0,0,1}
\providecolor{lyxdeleted}{rgb}{1,0,0}

 \let\oldforeign@language\foreign@language
 \DeclareRobustCommand{\foreign@language}[1]{%
   \lowercase{\oldforeign@language{#1}}}
\theoremstyle{plain}
\newtheorem{thm}{\protect\theoremname}
\theoremstyle{definition}
\newtheorem{defn}[thm]{\protect\definitionname}
\theoremstyle{plain}
\newtheorem{lem}[thm]{\protect\lemmaname}

\ifCLASSOPTIONcompsoc
\usepackage[caption=false,font=normalsize,labelfont=sf,textfont=sf]{subfig}
\else
\usepackage[caption=false,font=footnotesize]{subfig}
\fi

\usepackage{breakurl}
\usepackage{times}
\usepackage{epsfig}
\usepackage{multirow}
\usepackage{mdwlist}
\usepackage{algorithm}
\usepackage{algorithmic}%
\usepackage{amsmath}
\usepackage{graphicx} 
\usepackage{url}

\@ifundefined{showcaptionsetup}{}{%
 \PassOptionsToPackage{caption=false}{subfig}}
\usepackage{subfig}
\makeatother

\providecommand{\definitionname}{\inputencoding{latin9}Definition}
\providecommand{\lemmaname}{\inputencoding{latin9}Lemma}
\providecommand{\theoremname}{\inputencoding{latin9}Theorem}

\begin{document}
\global\long\def\mtbfM{\mathbf{M}}
 \global\long\def\mtbfm{\mathbf{m}}
 \global\long\def\mtbfY{\mathbf{Y}}
 \global\long\def\mtbfy{\mathbf{y}}

\global\long\def\mtbfA{\mathbf{A}}
 \global\long\def\mtbfa{\mathbf{a}}
 \global\long\def\mtbfE{\mathbf{E}}
 \global\long\def\mtbfe{\mathbf{e}}

\global\long\def\mtbfC{\mathbf{C}}
 \global\long\def\mtbfc{\mathbf{c}}
\global\long\def\mtbfU{\mathbf{U}}
 \global\long\def\mtbfu{\mathbf{u}}

\global\long\def\mtbfW{\mathbf{W}}
 \global\long\def\mtbfD{\mathbf{D}}
 \global\long\def\mtbfL{\mathbf{L}}
 \global\long\def\mtbfz{\mathbf{z}}

\global\long\def\mtbldM{\mathbf{M}}
 \global\long\def\mtbldm{\mathbf{m}}
 \global\long\def\mtbldY{\mathbf{Y}}
 \global\long\def\mtbldy{\mathbf{y}}

\global\long\def\mtbldA{\mathbf{A}}
 \global\long\def\mtblda{\mathbf{a}}
 \global\long\def\mtbldE{\mathbf{E}}
 \global\long\def\mtblde{\mathbf{e}}

\global\long\def\mtbldW{\mathbf{W}}
 \global\long\def\mtbldD{\mathbf{D}}
 \global\long\def\mtbldL{\mathbf{L}}
 \global\long\def\mtbldz{\mathbf{z}}

\global\long\def\tensorY{\mathcal{Y}}
 \global\long\def\tensorA{\mathcal{A}}
 \global\long\def\inSetsR{\in\mathbb{R}}

\title{Structured Sparse Method for Hyperspectral Unmixing}

\author{Feiyun~Zhu, Ying~Wang, Shiming~Xiang, Bin~Fan, and~Chunhong~Pan%
\thanks{Feiyun~Zhu, Ying~Wang, Shiming~Xiang, Bin~Fan and~Chunhong~Pan
are with the National Laboratory of Pattern Recognition, Institute
of Automation, Chinese Academy of Sciences (e-mail: \{fyzhu, ywang,
smxiang, bfan and chpan\}@nlpr.ia.ac.cn).%
}}

\markboth{ISPRS Journal of Photogrammetry and Remote Sensing}{F. Y. Zhu \MakeLowercase{\emph{et al.}}:
SS-NMF for Hyper-Unmixing}
\maketitle
\begin{abstract}
Hyperspectral Unmixing (HU) has received increasing attention in the
past decades due to its ability of unveiling information latent in
hyperspectral data. Unfortunately, most existing methods fail to take
advantage of the spatial information in data. To overcome this limitation,
we propose a Structured Sparse regularized Nonnegative Matrix Factorization
(SS-NMF) method from the following two aspects. First, we incorporate
a graph Laplacian to encode the manifold structures embedded in the
hyperspectral data space. In this way, the highly similar neighboring
pixels can be grouped together. Second, the lasso penalty is employed
in SS-NMF for the fact that pixels in the same manifold structure
are sparsely mixed by a common set of relevant bases. These two factors
act as a new structured sparse constraint. With this constraint, our
method can learn a compact space, where highly similar pixels are
grouped to share correlated sparse representations. Experiments on
real hyperspectral data sets with different noise levels demonstrate
that our method outperforms the state-of-the-art methods significantly. \end{abstract}

\section{Introduction}

Hyperspectral imaging has been widely used in many fields since it
provides the ability to record the same scene with hundreds of contiguous
and narrow spectral bands\ \cite{Keshava_03_LJ_unmixingSurvey,Tilts_12_ISPRS_Hyper}.
Due to the low spatial resolution of the sensor, the spectra of spatial
neighboring substances inevitably merge together, leading to mixed
pixels in hyperspectral data. This makes Hyperspectral Unmixing (HU)
an essential step for hyperspectral image analysis. Generally, the
task of HU is to decompose each pixel spectrum into a set of constituent
spectra (called \emph{endmembers}, such as tree, water, soil, etc.),
and their corresponding percentages (called \emph{abundances})\ \cite{Keshava_01_ISP_SpectralUnmixing,Dias_12_AEORS_HUoverview}.
In this paper, we focus on the unsupervised HU problem, for which
both \emph{endmembers} and \emph{abundances} are unknown, making HU
a very challenging problem.

In general, the linear HU methods can be roughly classified into two
categories: geometrical methods\ \cite{Jose_05_TGRS_Vca,Chang_06_ITGRS_EndmemberExt,Martin_12_SelectedTopics_SSPP}
and statistical ones\ \cite{Dobigeon_ISP_BaysianHU,Wang_06_ITGRS_ICAextEndmember,Jose_09_WHISPERS_SplittingAugmentedLag,Jose_12_TGRS_HUbyDirchletComponents}.
Geometrical methods assume that hyperspectral pixels are located within
a simplex, whose vertices correspond to the \emph{endmembers}. N-FINDR\ \cite{Michael_99_PSCIS_nFindr}
and Vertex Component Analysis (VCA)\ \cite{Jose_05_TGRS_Vca} are
two typical geometrical methods. The former treats the vertices of
a simplex with the maximum volume as \emph{endmembers}, while the
latter searches the \emph{endmembers} through projection. The above
two methods are both simple and fast. Unfortunately, they require
the existence of pure pixels for each \emph{endmember}, which is usually
unavailable in practice. Moreover, geometrical methods fail to consider
local structures latent in hyperspectral data, leading to an inaccurate
estimation of \emph{endmembers} and \emph{abundances}. 
\begin{figure}[t]
\centering{}\includegraphics[width=0.72\textwidth]{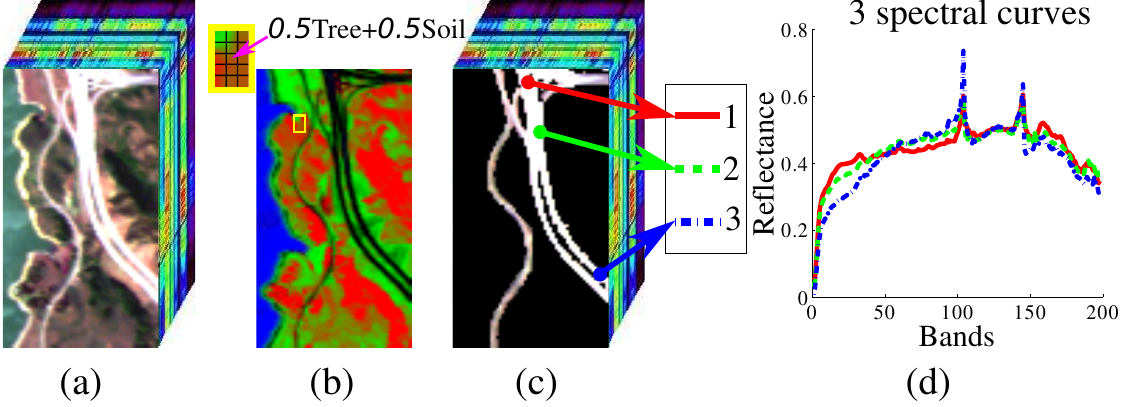}
\caption{Illustration of two priors. (a) Hyperspectral image. (b) \emph{Abundances}
of 4 targets, which are presented by proportions of 4 colors: Tree
(Red), Water (Blue), Soil (Green) and Road (Black). For example, there
are 15 mixed pixels in the yellow box patch; the color proportion
of the pixel pointed by arrow is \textit{0.5}Red (Tree)+\textit{0.5}Green
(Soil). (c) Road structure latent in data. (d) Spectral curves of
pixels-1, 2, 3 are similar in the road structure. (For interpretation
of the references to color in this figure legend, the reader is referred
to the web version of this article.) \label{fig:showOfObserv} }
\end{figure}

Accordingly, a number of statistical methods have been proposed. Nonnegative
Matrix Factorization (NMF)\ \cite{Lee_99_Nature_NMF} and its extensions
are the typical statistical methods. As an unsupervised method, NMF
tries to find two nonnegative matrices to approximate the original
matrix with their product. The nonnegative constraint on both factor
matrices is valuable for two reasons. First, the \emph{endmembers}
and their corresponding \emph{abundances} are both nonnegative, which
makes NMF physically suited to the HU problem. Second, the nonnegative
constraint allows additive, not subtractive, combinations, leading
to a parts-based representation \cite{Lee_99_Nature_NMF}. The parts-based
property makes representation results more intuitive and interpretable,
since psychological and physiological evidences have shown human brain
works in a parts-based way \cite{Palmer_77_CP_perceptualRepresent,Logothetis_96_ARN_VisObjRecognition}.

Although NMF has enjoyed a great success in many applications, such
as face analysis \cite{Stanzli_01_CVPR_locNMF,Roman_11_PAMI_EarthMoverNMF}
and documents clustering \cite{WeiXu_03_SIGIR_DocClusterNMF,Farial_06_InfProManag_DocumNMF},
there are three weaknesses. 
First, the solution space of NMF is large, which is caused by the
non-convex objective function. Many extensions have been proposed
to reduce the solution space by adding various constraints to the
NMF framework, such as non-smooth NMF \cite{Alberto_06_PAMI_nsNMF},
MVC-NMF \cite{Miao_07_ITGRS_NMFMVC}, MDMD-NMF \cite{Huck_10_WHISPERS_MDMD-NMF},
GL-NMF \cite{XLu_2013_TGRS_ManifoldSparseNMF} and so on. Second,
the parts-based property of NMF is usually not strong enough for the
HU problem, resulting in a less expressive estimation of \emph{endmembers}.
A possible solution is to employ a sparse constraint to the NMF framework
\cite{Plaza_11_ITGRS_SparseUnmixingHU,Qian_11_TGRS_NMF+l1/2}, with
a regularization parameter to control the weight of parts-based property.
Third, NMF method  does not consider similarities between neighboring
pixels, leading to structure information unused. 

In this paper we propose an effective method, named Structured Sparse
NMF (SS-NMF), to overcome the above three limitations of NMF. The
motivation is based on two priors. First, due to the smoothness of
local patches in images, there are many regions where the spectral
curves of all pixels should be similar with each other. Fig.~\ref{fig:showOfObserv}c
illustrates an example, where the pixels in the road region have similar
visual appearance (see Fig.~\ref{fig:showOfObserv}d). These regions
are the structures that should be preserved while unmixing. Second,
most pixels in hyperspectral data are mixed by only a few \emph{endmembers}
\cite{Qian_11_TGRS_NMF+l1/2,LiuXueSong_2011_TGRS_ConstrainedNMF,Plaza_11_ITGRS_SparseUnmixingHU}.
In summary, the pixels in the same structure are sparsely mixed by
a common set of relevant \emph{endmembers}. This can be obviously
observed from Fig.~\ref{fig:showOfObserv}b, where the proportions
of four colors represent the \emph{abundances} of four \emph{endmembers}:
Red (Tree), Blue (Water), Green (Soil) and Black (Road). As can be
seen, the colors in Fig.~\ref{fig:showOfObserv}b are regional smooth
and sparsely mixed by Red, Blue, Green and Black ink.

Based on the above analysis, we introduce the Structured Sparse constraint
(SS-constraint in short) from the following two aspects. First, we
model the latent structures by encoding a weighted graph on the hyperspectral
pixels. 
In this way, the highly related neighboring pixels are grouped on
the graph. Accordingly, a graph constraint is employed in SS-NMF to
transfer these structures into the unmixing results. Second, we use
the lasso penalty to sparsely encode the \emph{abundances} for each
pixel. With the SS-constraint, we can learn an \emph{abundance} space,
in which \emph{abundances} are regional smooth and sparse, as shown
in Fig.~\ref{fig:showOfObserv}b. Experiments on real hyperspectral
data with different noise levels illustrate that SS-NMF outperforms
the state-of-the-art methods.

Below are some remarks about our method: 
\begin{itemize}
\item We propose a meaningful SS-constraint and apply it in the NMF framework.
The basic idea is to encourage highly similar pixels on the graph
to share correlated sparse \emph{abundances}, which is similar with
\cite{Chen_10_arXiv_GraphedSparse}. With the help of SS-constraint,
SS-NMF overcomes the three limitations of NMF for the HU problem.
\item SS-NMF is an effective method for HU problems. We compare our method
with several state-of-the-art methods on two hyperspectral data. Both
quantitative and qualitative performances show that our method outperforms
the state-of-the-art methods significantly.
\end{itemize}
The remainder of this paper is organized as follows. In Section 2,
we briefly review Linear Mixture Model (LMM) and Nonnegative Matrix
Factorization (NMF) method. Section 3 presents the formulation, the
optimization algorithm, the convergence proof and the analyzes of
computation complexity for SS-NMF method. Extensive results as well
as detailed comparisons and analyses are presented in Section 4, followed
by conclusions in Section 5.

\section{Preliminaries}

In this section, we introduce the Linear Mixture Model (LMM) and the
NMF method. The former is the fundamental model for the methods mentioned
in the following sections, while the latter is the basic method we
build on.

\subsection{Linear Mixture Model (LMM)}

LMM is the most popular model for hyperspectral image analysis \cite{Geng_13_ISPRS_endmemberExtr,Keshava_01_ISP_SpectralUnmixing}.
Suppose we are given a hyperspectral image with $L$ bands and $N$
pixels $\left\{ \mtbfy_{n}\right\} _{n=1}^{N}\inSetsR_{+}^{L}$. LMM
assumes that each pixel $\mtbldy$ can be approximated by a nonnegative
linear combination of $K$ \emph{endmembers} (bases) as follows 
\begin{equation}
\mtbldy=\mtbldM\mtblda+\mtblde,\label{eq:LinearUnmixingModelInVector}
\end{equation}
where $\mtbldM\equiv\left[\mtbldm_{1},\cdots,\mtbldm_{K}\right]\inSetsR_{+}^{L\times K}$
is the \emph{endmember} matrix with $K$ \emph{endmembers}, $\mtblda\inSetsR_{+}^{K}$
is the \emph{abundance} vector associated with pixel $\mtbldy$, $\mtblde\inSetsR^{L}$
is a residual term. For all pixels together, we have the matrix form
of LMM as 
\begin{equation}
\mtbldY=\mtbldM\mtbldA+\mtbldE,\label{eq:LinearUnmixingModelInMatrix}
\end{equation}
where $\mtbldY\equiv\left[\mtbldy_{1},\cdots,\mtbldy_{N}\right]\inSetsR_{+}^{L\times N}$
is a hyperspectral image with column vectors corresponding to pixels,
$\mtbldA\equiv\left[\mtblda_{1},\cdots,\mtblda_{N}\right]\inSetsR_{+}^{K\times N}$
is the \emph{abundance} matrix, whose column vector $\mtblda_{n}\equiv\left[A_{1n},\cdots,A_{Kn}\right]^{T}\inSetsR_{+}^{K}$
contains fractional \emph{abundances of }$K$ \emph{endmembers} for
pixel $\mtbldy_{n}$. $\mtbldz_{k}\equiv[A_{k1},\cdots,A_{kN}]\inSetsR_{+}^{1\times N}$
is the $k$-th row vector in $\mtbfA$ that contains fractional \emph{abundances}
of the $k$-th \emph{endmember} (i.e. $\mtbldm_{k}$) associated with
all pixels. Here $\mtbldz_{k}$ is called the $k$-th \emph{abundance}
map. $\mtbldE\inSetsR^{L\times N}$ is a residual term.

For HU problems, we have $K\ll\mbox{min}\left(L,N\right)$, which
intimates very few \emph{endmembers} are latent in the hyperspectral
data space. In addition, most pixels are mixed by parts of the $K$
\emph{endmembers}. Therefore, good approximations can be achieved
if the following two conditions are satisfied: (1) they can unveil
intrinsic structures in hyperspectral data \cite{Lee_99_Nature_NMF},
whose rank is $K$; (2) the learned \emph{endmembers} are expressive
\cite{Stanzli_01_CVPR_locNMF}.

\subsection{Nonnegative Matrix Factorization}

NMF is a popular method that builds in terms of LMM. The aim is to
approximately decompose a large nonnegative matrix $\mtbldY$ into
two nonnegative matrices $\mtbldM$ and $\mtbldA$ by minimizing the
following objective function \cite{Lee_99_Nature_NMF}, 
\begin{equation}
\min_{\mtbldM,\mtbldA}\ \frac{1}{2}\|\mtbldY-\mtbldM\mtbldA\|_{F}^{2}\quad\mbox{s.t.}\ \ \mtbldM\geq0,\mtbldA\geq0,\label{eq:NMFobjectiveFunction}
\end{equation}
where $\|\cdot\|_{F}$ is the \emph{Frobenius }norm, all elements
in $\mtbldM\inSetsR_{+}^{L\times K}$ and $\mtbldA\inSetsR_{+}^{K\times N}$
are unknown and nonnegative. Although \eqref{eq:NMFobjectiveFunction}
is convex with respect to $\mtbldM$ and $\mtbldA$ respectively,
it is non-convex for both variables together \cite{Lee_00_NIPS_NMF}.
Therefore it is unrealistic to find global minima. Alternately, \cite{Lee_00_NIPS_NMF}
have proposed the multiplicative update rules as follows 
\begin{equation}
M_{lk}\leftarrow M_{lk}\frac{\left(\mtbfY\mtbfA^{T}\right)_{lk}}{\left(\mtbfM\mtbfA\mtbfA^{T}\right)_{lk}},\qquad A_{kn}\leftarrow A_{kn}\frac{\left(\mtbfM^{T}\mtbfY\right)_{kn}}{\left(\mtbfM^{T}\mtbfM\mtbfA\right)_{kn}},
\end{equation}
which have been proved to be non-increasing. There are other methods
to solve the problem \eqref{eq:NMFobjectiveFunction}, such as the
active-set \cite{Kim_08_JMAA_activeSetNMF} method, the alternation
nonnegative least least squares \cite{Berry_07_CSDA_AlgorithNMF},
the projected gradient descent \cite{Lin_07_NC_ProjecGradMethodNMF}
and so on. 
\begin{figure}[t]
\centering{}\includegraphics[width=0.72\textwidth]{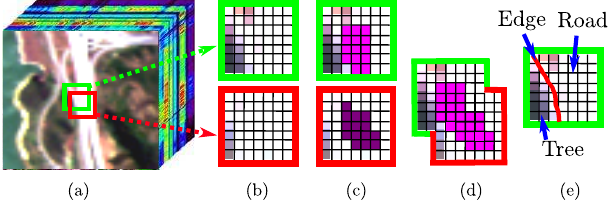}
\caption{Highly similar neighboring pixels could be grouped by local structures.
(a) A hyperspectral image. (b) Two local windows in (a). (c) The highly
similar neighboring pixels are grouped together. (d) When the local
structures are combined, the highly similar pixels in the two windows
are grouped together. (e) Window covers an edge, where `Tree' and
`Road' are on the two sides. \label{fig:localGraph} }
\end{figure}

Compared with the Vector Quantization (VQ) \cite{Gersho_91_KAP_VQS}
method and the Principal Component Analysis (PCA) \cite{Jolliffe_02_Springer_Pca}
method, the nonnegative constraint on both factor matrices enables
a parts-based representation, which makes NMF superior to PCA and
VQ in facial image analyses and document representations \cite{Lee_99_Nature_NMF}.

\section{Structured Sparse NMF (SS-NMF)\label{section:SS-NMF_formulation} }

As mentioned before, NMF applies the nonnegative constraint on the
factor matrices, which leads to a parts-based representation of the
original data. However, it fails to take consideration of the sparse
prior on the \emph{abundances} and the manifold structures hidden
in hyperspectral data, which are essential for the HU problem. In
this section, we introduce the SS-NMF method to overcome these limitations
by adopting a structured sparse constraint.

\subsection{Formulation for SS-NMF \label{sub:Formulation-forSS-NMF}}

In the framework of SS-NMF, the fundamental problem is how to construct
the structures. The manifold learning theory \cite{Belkin_01_NIPS_LaplacianEigenmap}
and spectral graph theory \cite{Chung_97_AMS_graphSpectralTheory,smXiang_2010_PRMI_LocSplineRegression}
provide us with a good idea that local structures can be learned by
encoding similarities between pixels on the $p$ nearest neighbor
graph.  Combining the local overlapping structures, we can get the
graph structure that contains full similarity information between
pixel pairs.

Typically, there are two ways to define the $p$ nearest neighbors:
considering the feature distances only and the spatial distances only.
The former is well suited for less geometrically intuitive databases
that treat each image as a node, such as the Yale database %
\footnote{http://cvc.yale.edu/projects/yalefaces/yalefaces.html%
} and the ORL database %
\footnote{http://www.uk.research.att.com/facedatabase.html%
} \cite{Belkin_01_NIPS_LaplacianEigenmap}; While the latter is adapted
to geometrically motivated databases that treat each pixel, which
are on the grids of a 2-D plane, as nodes, such as the hyperspectral
images. Due to the local smoothness prior of the hyperspectral images,
the spatially neighboring pixels are very likely to enjoy similar
features (i.e. spectral signatures). However, this prior does not
hold for image edges, where pixels on the two sides are dissimilar.
This motivates the weighting method for the proposed SS-NMF, which
would be thoroughly introduced in the following paragraph.

The proposed weighting method considers the spatial distance and the
feature distance simultaneously when constructing the local structures.
Suppose we are given a hyperspectral image $\mtbldY\inSetsR_{+}^{L\times N}$
with $N$ pixels. We view each pixel as one node and construct a weighted
graph on the $N$ nodes. The weight between nodes $\mtbldy_{i}$ and
$\mtbldy_{j}$ is obtained by the Spectral Angle Distance (SAD) metric
\cite{Jia_09_TGRS_ConstainedNMF,Jose_05_TGRS_Vca} defined as follows,
\begin{equation}
W_{ij}=\begin{cases}
\text{SAD}\left(\mtbldy_{i},\mtbldy_{j}\right), & \mtbldy_{j}\in\mathcal{N}(\mtbldy_{i})\ \text{or}\ \mtbldy_{i}\in\mathcal{N}(\mtbldy_{j})\\
\qquad0, & \mbox{otherwise}
\end{cases}\label{eq:neighborWeight}
\end{equation}
where 
\begin{equation}
\text{SAD}\left(\mtbldy_{i},\mtbldy_{j}\right)=\arccos\left(\frac{\mtbldy_{i}^{T}\mtbldy_{j}}{\|\mtbldy_{i}\|\cdot\|\mtbldy_{j}\|}\right).\label{eq:sadMetric}
\end{equation}
In \eqref{eq:neighborWeight}, the neighbors of $\mtbldy_{i}$ are
collected in $\mathcal{N}(\mtbldy_{i})$, and there are two conditions
for pixel $\mtbldy_{j}\in\mathcal{N}(\mtbldy_{i})$: (1) nearest spatial
distance, i.e., $\mtbldy_{j}$ is in the $m\times m$ ($m=7$ in this
paper) local window of $\mtbldy_{i}$ (see Fig.~\ref{fig:localGraph}b);
(2) nearest feature distance, i.e., calculating the SAD similarities
between $\mtbldy_{i}$ and its neighboring pixels in the $m\times m$
local window, $\text{SAD}(\mtbldy_{i},\mtbldy_{j})$ is among the
top 30\% (empirical threshold), as shown in Fig.~\ref{fig:localGraph}c.
As a result, we can partly avoid spreading the graph across dissimilar
pixels. This is essential when the local window covers the edges,
where pixels on the two sides belonging to different targets. Fig.~\ref{fig:localGraph}e
illustrates an example, where `Road' is on one side of the edge, while
`Tree' is on the other side. The SAD metric, defined in Eq.~\eqref{eq:sadMetric},
is a suitable metric, since it is an essential estimation metric for
\emph{endmembers} \cite{Qian_11_TGRS_NMF+l1/2,Jia_09_TGRS_ConstainedNMF,Jose_05_TGRS_Vca,LiuXueSong_2011_TGRS_ConstrainedNMF}.

Through the above process, we can build the local weighted graph,
or structure, with node $\mtbldy_{i}$ at the center. In this way,
the highly similar neighboring pixels are grouped together (see Figs.~\ref{fig:localGraph}c
and~\ref{fig:localGraph}d). Repeating the above procedures for each
pixel, we can construct the weight matrix $\mtbldW$ that contains
the manifold structures latent in hyperspectral data. Our goal is
to transfer these structures to the learned \emph{abundance} space.
This could be done by first measuring the distance between \emph{abundance}
vectors $\mtblda_{i}$ and $\mtblda_{j}$
\begin{equation}
d\left(\mtblda_{i},\mtblda_{j}\right)=\|\mtblda_{i}-\mtblda_{j}\|_{2}^{2},
\end{equation}
where $\|\cdot\|_{2}$ denotes the $\ell_{2}$-norm of a vector, measuring
the Euclidean distance. One intuitive way is to make $d\left(\mtblda_{i},\mtblda_{j}\right)$
small, if $\mtbldy_{i}$ and $\mtbldy_{j}$ are similar \cite{Belkin_01_NIPS_LaplacianEigenmap}.
This can be achieved by 
\begin{align}
\mathcal{R} & =\sum_{i=1}^{N}\sum_{j=1}^{N}d\left(\mtblda_{i},\mtblda_{j}\right)W_{ij}=\mbox{Tr}\left(\mtbldA\mtbldD\mtbldA^{T}\right)-\mbox{Tr}\left(\mtbldA\mtbldW\mtbldA^{T}\right)=\mbox{Tr}\left(\mtbldA\mtbldL\mtbldA^{T}\right),\label{eq:laplacianGraphConstraint}
\end{align}
where $\mbox{Tr}\left(\cdot\right)$ represents the trace of a matrix
and $\mtbldD$ is a diagonal matrix whose elements are column (or
row, as $\mtbldW$ is symmetric) sums of $\mtbldW$, i.e., $D{}_{ii}=\sum_{j=1}^{N}W_{ji}$.
$\mtbldL=\mtbldD-\mtbldW$ is called graph Laplacian \cite{Chung_97_AMS_graphSpectralTheory,Cai_11_PAMI_GNMF}.

By minimizing Eq.~\eqref{eq:laplacianGraphConstraint}, the manifold
structures in the original data space are transferred to the learned
\emph{abundance} space. However, Eq.~\eqref{eq:laplacianGraphConstraint}
has not considered the fact that most manifold structures in hyperspectral
data are dominated by specific sets of only a few \emph{endmembers}.
Here a lasso penalty \cite{Tibshirani_94_Statist_Lasso,Donoho_96_ITIT_CS}
on the \emph{abundance} matrix is adopted, resulting in a structured
sparse constraint as 
\begin{align}
\mathcal{R} & =\frac{\lambda}{2}\sum_{i=1}^{N}\sum_{j=1}^{N}\mtbldW_{ij}\|\mtblda_{i}-\mtblda_{j}\|_{2}^{2}+\alpha\sum_{n=1}^{N}\|\mtblda_{n}\|_{1}\label{eq:ss_constraint_1}\\
 & =\frac{\lambda}{2}\mbox{Tr}\left(\mtbldA\mtbldL\mtbldA^{T}\right)+\alpha\|\mtbldA\|_{1},\label{eq:ss_constraint_2}
\end{align}
where $\|\mtbldA\|_{1}=\sum_{n=1}^{N}\sum_{k=1}^{K}|A_{kn}|$ is used
to achieve an expressive set of \emph{endmembers} \cite{Stanzli_01_CVPR_locNMF}.
Note that the structure regularization defined in Eq.~\eqref{eq:laplacianGraphConstraint}
provides the lasso penalty with a structure constraint, under which
the sparse \emph{abundance }vectors related to the same structure
tend to be similar. For example, $\mtbldy_{i}$ and $\mtbldy_{j}$
are in the same local structure (i.e., $W_{ij}$ is big), then their
corresponding sparse \emph{abundances} $\mtblda_{i}$ and $\mtblda_{j}$
tend to be similar. In addition, $W_{ij}$ controls how strongly $\mtblda_{i}$
and $\mtblda_{j}$ share similar \emph{abundances}. This renders the
structured sparse constraint defined in Eq.~\eqref{eq:ss_constraint_2}.

Given a hyperspectral data $\mtbldY\inSetsR_{+}^{L\times N}$, similar
to the NMF method, the objective function of SS-NMF is defined as
follows 
\begin{equation}
\begin{aligned}\mathcal{O}=\frac{1}{2}\|\mtbldY-\mtbldM\mtbldA\|_{F}^{2}+\frac{\lambda}{2}\mbox{Tr}\left(\mtbldA\mtbldL\mtbldA^{T}\right)+\alpha\|\mtbldA\|_{1},\end{aligned}
\label{eq:objFunctionOfOurAlg}
\end{equation}
where $\lambda\geq0,\alpha\geq0$ are regularization parameters that
control the complexity of the model, all elements in $\mtbldM$, $\mtbldA$
are nonnegative and unknown. We will present the optimization method
to solve the problem defined in Eq.~\eqref{eq:objFunctionOfOurAlg}
in the following subsection.

\subsection{Optimization for SS-NMF }

Similar to the NMF problem, the objective function~\eqref{eq:objFunctionOfOurAlg}
is non-convex for $\mtbldM$ and $\mtbldA$ together. An iterative
algorithm, which could reach local minima, is introduced in this subsection.
When considered the nonnegative constraint on $\mtbldM$ and $\mtbldA$,
the objective function \eqref{eq:objFunctionOfOurAlg} could be reformulated
as
\begin{align}
\min_{\mtbldM,\mtbldA}\  & \frac{1}{2}\|\mtbldY-\mtbldM\mtbldA\|_{F}^{2}+\frac{\lambda}{2}\mbox{Tr}\left(\mtbldA\mtbldL\mtbldA^{T}\right)+\alpha\sum_{k=1}^{K}\sum_{n=1}^{N}A_{kn},\quad\mbox{s.t.}\ \ \mtbldM\geq0,\mtbldA\geq0.\label{eq:objFuncOfOurAlgNew}
\end{align}
Let $\theta_{lk}$, $\phi_{kn}$ be the Laplacian multipliers for
constraint $M_{lk}\geq0$ and $A_{kn}\geq0$ respectively, and $\Theta=[\theta_{lk}]\inSetsR_{+}^{L\times K}$,
$\Phi=[\phi_{kn}]\inSetsR_{+}^{K\times N}$. The Lagrange $\mathcal{L}$
is given by
\begin{equation}
\mathcal{L}=\frac{1}{2}\|\mtbldY-\mtbldM\mtbldA\|_{F}^{2}+\alpha\sum_{k=1}^{K}\sum_{n=1}^{N}A_{kn}+\frac{\lambda}{2}\mbox{Tr}\left(\mtbldA\mtbldL\mtbldA^{T}\right)+\mbox{Tr}\left(\Theta\mtbldM^{T}\right)++\mbox{Tr}\left(\Phi\mtbldA^{T}\right).
\end{equation}
 We can further obtain the partial derivative of $\mathcal{L}$ with
respect to $\mtbfM$ and $\mtbfA$ as 
\begin{align}
\frac{\partial\mathcal{L}}{\partial\mtbldM} & =\mtbldM\mtbldA\mtbldA^{T}-\mtbldY\mtbldA^{T}+\Theta\label{eq:partialDerive_M}
\end{align}
\begin{equation}
\frac{\partial\mathcal{L}}{\partial\mtbldA}=\mtbldM^{T}\mtbldM\mtbldA-\mtbldM^{T}\mtbldY+\lambda\mtbldA\mtbldL+\alpha+\Phi.\label{eq:partialDerive_A}
\end{equation}
Based on the Karush-Kuhn-Tucker conditions $\theta_{lk}M_{lk}=0$
and $\phi_{kn}A_{kn}=0$, we could obtain the following equations
by letting the above partial derivatives equal to zero and multiplying
both sides with $M_{lk}$ and $A_{kn}$ for Eqs.~\eqref{eq:partialDerive_M}
and~\eqref{eq:partialDerive_A} respectively,

\begin{align}
\left(\mtbldM\mtbldA\mtbldA^{T}\right)_{lk}M_{lk}-\left(\mtbldY\mtbldA^{T}\right)_{lk}M_{lk} & =0\label{eq:partDeriveEquaM}\\
\left(\mtbldM^{T}\mtbldM\mtbldA\right)_{kn}A_{kn}-\left(\mtbldM^{T}\mtbldY\right)_{kn}A_{kn}+\lambda\left(\mtbldA\mtbldL\right)_{kn}A_{kn}+\alpha A_{kn} & =0.\label{eq:partDeriveEquaA}
\end{align}
When the equation $\mtbfL=\mtbfD-\mtbfW$ mentioned before is considered,
Eq.~\eqref{eq:partDeriveEquaA} could be rewritten as
\begin{equation}
\left[\left(\mtbldM^{T}\mtbldM\mtbldA\right)_{kn}+\lambda\left(\mtbldL\mtbfD\right)_{kn}+\alpha\right]A_{kn}=\left[\left(\mtbldM^{T}\mtbldY\right)_{kn}+\lambda\left(\mtbldL\mtbfW\right)_{kn}\right]A_{kn}.\label{eq:partDeriveEquaA_2}
\end{equation}
Solving Eqs.~\eqref{eq:partDeriveEquaM} and \eqref{eq:partDeriveEquaA_2},
we have the updating rules as 
\begin{equation}
M_{lk}\leftarrow M_{lk}\frac{\left(\mtbldY\mtbldA^{T}\right)_{lk}}{\left(\mtbldM\mtbldA\mtbldA^{T}\right)_{lk}}\label{eq:updatingRule_M}
\end{equation}

\begin{equation}
A_{kn}\leftarrow A_{kn}\frac{\left(\mtbldM^{T}\mtbldY+\lambda\mtbldA\mtbldW\right)_{kn}}{\left(\mtbldM^{T}\mtbldM\mtbldA+\lambda\mtbldA\mtbldD+\alpha\right)_{kn}}.\label{eq:updatingRule_A}
\end{equation}
Therefore, Eq.~\eqref{eq:objFuncOfOurAlgNew} could be solved by
alternately updating $\mtbldM$ and $\mtbldA$ according to Eqs.~\eqref{eq:updatingRule_M}
and \eqref{eq:updatingRule_A} respectively. Our algorithm of SS-NMF
is summarized in Algorithm~\ref{alg:SS-NMF}.

It is worthwhile to point out that if $\mtbfM$ and $\mtbfA$ form
the solution for NMF algorithm, then $\mtbfD\mtbfU$ and $\mtbfU^{-1}\mtbfA$
are the solution for any diagonal matrix $\mtbfU$ with positive diagonal
elements. To get rid of this uncertainty, a simple and widely used
method is to scale each column of $\mtbfM$ or each row of $\mtbfA$
to be unit $\ell_{2}$-norm \cite{WeiXu_03_SIGIR_DocClusterNMF} (or
$\ell_{1}$-norm). This can be achieved by,
\begin{equation}
M_{lk}\leftarrow\frac{M_{lk}}{\sqrt{\sum_{l=1}^{L}M_{lk}^{2}}},\qquad A_{kn}\leftarrow A_{kn}\sqrt{\sum_{l}M_{lk}^{2}}.\label{eq:eleminatingUncertain}
\end{equation}
The proposed SS-NMF also employs this strategy to eliminate the uncertainty,
which is essential in the sense of computer realization. 
\begin{algorithm}[htp]
\caption{\textbf{for SS-NMF \label{alg:SS-NMF}}}

\textbf{Input:} Hyperspectral data $\mtbldY$, the number of \emph{endmembers}
$K$, parameters $\lambda$, $\alpha$, and threshold $\tau$. 

\textbf{Output:} \emph{Endmember} matrix $\mtbldM\inSetsR_{+}^{L\times K}$
and \emph{abundance} matrix $\mtbldA\inSetsR_{+}^{K\times N}$ 

\begin{algorithmic}[1] 

\STATE \textbf{Initialize} the\emph{ endmember} matrix by randomly
selecting $K$ pixels from $\mtbldY$. Each newly selected pixel should
be dissimilar with the pixels have been selected \cite{LiuXueSong_2011_TGRS_ConstrainedNMF}.
Initialize $\mtbldA$ with random positive values, and scale each
column of $\mtbldA$ to be unit $\ell_{1}$-norm. 

\STATE Construct the weight matrix $\mtbldW$ on the hyperspectral
image based on Eqs.~\eqref{eq:neighborWeight}, \eqref{eq:sadMetric}
and calculate the degree matrix $\mtbldD$ and graph laplacian matrix
$\mtbldL$. 

\WHILE {$\frac{\|\mathcal{O}_{old}-\mathcal{O}_{new}\|}{\mathcal{O}_{old}}\geq\tau$
($\mathcal{O}$ defined in Eq.~\eqref{eq:objFunctionOfOurAlg})} 

\STATE update $\mtbldA$ by Eq.~\eqref{eq:updatingRule_A}.

\STATE update $\mtbldM$ by Eq.~\eqref{eq:updatingRule_M}.

\STATE scale $\mtbfM$ and $\mtbfA$ by Eq.~\eqref{eq:eleminatingUncertain}
after each iteration.

\ENDWHILE \\
\STATE Output $\mtbldM$ and $\mtbldA$ as the final unmixing result. 

\end{algorithmic} \label{code:recentEnd} 
\end{algorithm}

\subsection{Proof of Convergence }

In this subsection, we demonstrate that the optimization problem Eq.~\eqref{eq:objFuncOfOurAlgNew}
is non-increasing by using the updating rules~\eqref{eq:updatingRule_M}
and~\eqref{eq:updatingRule_A} during each iteration, and it finally
converges to local minima after finite iterations. Since the updating
rule for $\mtbfM$ is the same as that of NMF  and the convergence
proof is provided by \cite{Lee_00_NIPS_NMF}, we only focus on the
proof of the updating rule~\eqref{eq:updatingRule_A} for $\mtbfA$.
A common skill used in the Expectation Maximization (EM) algorithm
\cite{Dempster_77_JRSS_EM,Saul_97_EMNLP_mixedOrderMarkov} is adopted
by introducing an auxiliary function, which is exactly an upper bound
function.
\begin{defn}
$G\left(\mtbfA,\mtbfA'\right)$ is an auxiliary function for $F\left(\mtbfA\right)$
if the following properties are satisfied: 
\[
G\left(\mtbfA,\mtbfA'\right)\geq F\left(\mtbfA\right),\qquad G\left(\mtbfA,\mtbfA\right)=F\left(\mtbfA\right).
\]
 
\end{defn}
By minimizing the energy of $G\left(\mtbfA,\mtbfA'\right)$ exactly
given by 
\begin{equation}
\mtbfA^{\left(t+1\right)}=\arg\min_{\mtbfA}G\left(\mtbfA,\mtbfA^{\left(t\right)}\right),\label{eq:updatingByAuxiliaryFunc}
\end{equation}
we can get a suitable solution $\mtbfA^{\left(t+1\right)}$ that makes
$F\left(\mtbfA\right)$ non-increasing 
\begin{equation}
F\left(\mtbfA^{\left(t+1\right)}\right)\leq F\left(\mtbfA^{\left(t\right)}\right).
\end{equation}

\begin{IEEEproof}
This is because of the following inequalities,
\begin{equation}
F\left(\mtbfA^{\left(t+1\right)}\right)=G\left(\mtbfA^{\left(t+1\right)},\mtbfA^{\left(t+1\right)}\right)\leq G\left(\mtbfA^{\left(t+1\right)},\mtbfA^{\left(t\right)}\right)\leq G\left(\mtbfA^{\left(t\right)},\mtbfA^{\left(t\right)}\right)=F\left(\mtbfA^{\left(t\right)}\right).
\end{equation}

\end{IEEEproof}
Now, the objective function defined in Eq.~\eqref{eq:objFunctionOfOurAlg}
is only related to variable $\mtbfA$. It could be represented by
$F\left(\mtbfA\right)$ as 
\begin{equation}
F\left(\mtbfA\right)=\frac{1}{2}\|\mtbldY-\mtbldM\mtbldA\|_{F}^{2}+\frac{\lambda}{2}\mbox{Tr}\left(\mtbldA\mtbldL\mtbldA^{T}\right)+\alpha\|\mtbldA\|_{1}.\label{eq:F_func_A}
\end{equation}
Specifically, it is a quadratic function that equals to the Taylor
expansion
\begin{eqnarray}
F(\mtbldA) & = & F(\mtbldA^{\left(t\right)})+\text{Tr}\left(\mtbfC^{T}\nabla F(\mtbldA^{\left(t\right)})\right)+\frac{1}{2}\text{Tr}\left(\mtbfC^{T}\left(\mtbldM^{T}\mtbldM\right)\mtbfC\right)+\text{\ensuremath{\frac{\lambda}{2}}}\text{Tr}\left(\mtbfC\mtbldL\mtbfC{}^{T}\right),\label{eq:objTaylorExpan}
\end{eqnarray}
where $\mtbfC=\left(\mtbldA-\mtbldA^{\left(t\right)}\right)\inSetsR^{K\times N}$.
A function constituted based on the updating rule~\eqref{eq:updatingRule_A}
is given by
\begin{equation}
G\left(\mtbfA,\mtbfA^{\left(t\right)}\right)=F(\mtbldA^{\left(t\right)})+\text{Tr}\left(\mtbfC^{T}\nabla F(\mtbldA^{\left(t\right)})\right)+\frac{1}{2}\sum_{k=1}^{K}\sum_{n=1}^{N}Q_{kn}C_{kn}^{2},\label{eq:auxiliaryFunc}
\end{equation}
where 
\begin{equation}
Q_{kn}=\frac{\left(\mtbldM^{T}\mtbldM\mtbldA^{\left(t\right)}+\lambda\mtbldA^{\left(t\right)}\mtbldD+\alpha\right)_{kn}}{A_{kn}^{\left(t\right)}}.\label{eq:auxQuadraticTerm}
\end{equation}
This means that we could obtain the updating rule~\eqref{eq:updatingRule_A}
by minimizing the function $G\left(\mtbfA,\mtbfA^{\left(t\right)}\right)$
at the $t$-th iteration step.
\begin{lem}
\label{lem:proofUpperFunct}The Function $G\left(\mtbfA,\mtbfA^{\left(t\right)}\right)$\textup{
defined in} \eqref{eq:auxiliaryFunc} is an auxiliary function (i.e.
upper bound function) for the function $F\left(\mtbfA\right)$ defined
in \eqref{eq:F_func_A}.\end{lem}
\begin{IEEEproof}
When $\mtbfA=\mtbfA^{\left(t\right)}$, i.e. $\mtbfC=\mathbf{0}$,
is satisfied, we have $F\left(\mtbfA\right)=F\left(\mtbldA^{\left(t\right)}\right)=G\left(\mtbfA,\mtbfA^{\left(t\right)}\right)$.
Specifically, when $\mtbfA\neq\mtbfA^{\left(t\right)}$ (i.e. $\mtbfC\neq\mathbf{0}$),
we have to prove $G\left(\mtbfA,\mtbfA^{\left(t\right)}\right)\text{\ensuremath{\geq}}F\left(\mtbfA\right)$.
Since the constant term and linear term in Eq.\eqref{eq:objTaylorExpan}
are equal to their counterparts in Eq.\eqref{eq:auxiliaryFunc}, Lemma~\ref{lem:proofUpperFunct}
could be proven by comparing the quadratic terms as 
\begin{equation}
\sum_{k=1}^{K}\sum_{n=1}^{N}Q_{kn}C_{kn}^{2}\geq\text{Tr}\left(\mtbfC^{T}\left(\mtbldM^{T}\mtbldM\right)\mtbfC\right)+\lambda\mbox{Tr}\left(\mtbfC\mtbldL\mtbfC^{T}\right).\label{eq:compareAuxAndTaylor}
\end{equation}
Since $\mtbfD$ is a diagonal matrix, Eq.~\eqref{eq:auxQuadraticTerm}
is simplified as
\begin{equation}
Q_{kn}=\frac{\left(\mtbldM^{T}\mtbldM\mtbldA^{\left(t\right)}\right)_{kn}}{A_{kn}^{\left(t\right)}}+\lambda D_{nn}+\frac{\alpha}{A_{kn}^{\left(t\right)}}.\label{eq:auxQuadraticTermNew}
\end{equation}
The inequality \eqref{eq:compareAuxAndTaylor} could be rewritten
as
\begin{equation}
\underset{\text{first term}}{\underbrace{\left\{ \sum_{k,n}\frac{\left(\mtbldM^{T}\mtbldM\mtbldA^{\left(t\right)}\right)_{kn}}{A_{kn}^{\left(t\right)}}C_{kn}^{2}-\text{Tr}\left(\mtbfC^{T}\mtbldM^{T}\mtbldM\mtbfC\right)\right\} }}+\underset{\text{second term}}{\underbrace{\left\{ \sum_{k,n}\left(\lambda D_{nn}+\frac{\alpha}{A_{kn}^{\left(t\right)}}\right)C_{kn}^{2}-\lambda\mbox{Tr}\left(\mtbfC\mtbldL\mtbfC^{T}\right)\right\} }\geq0,}\label{eq:compareAuxAndTaylor_v2}
\end{equation}
which could be proven by proving both terms to be greater than or
equal to zero. 

To begin with, the first term in Eq.~\eqref{eq:compareAuxAndTaylor_v2}
is compared as
\begin{align}
f_{1} & =\sum_{k=1}^{K}\sum_{n=1}^{N}\frac{\left(\mtbldM^{T}\mtbldM\mtbldA^{\left(t\right)}\right)_{kn}}{A_{kn}^{\left(t\right)}}C_{kn}^{2}-\text{Tr}\left(\mtbfC^{T}\mtbldM^{T}\mtbldM\mtbfC\right)\nonumber \\
 & =\sum_{k=1}^{K}\sum_{n=1}^{N}\frac{\left(\mtbldM^{T}\mtbldM\mtbldA^{\left(t\right)}\right)_{kn}}{A_{kn}^{\left(t\right)}}C_{kn}^{2}-\sum_{k=1}^{K}\sum_{l=1}^{K}\sum_{n=1}^{N}C_{kn}C_{ln}\left(\mtbfM^{T}\mtbfM\right)_{lk}\nonumber \\
 & =\frac{1}{2}\sum_{k=1}^{K}\sum_{n=1}^{N}\sum_{l=1}^{K}\left(\mtbldM^{T}\mtbldM\right)_{kl}\frac{\left(A_{ln}^{\left(t\right)}\right){}^{2}C_{kn}^{2}+\left(A_{kn}^{\left(t\right)}\right){}^{2}C_{ln}^{2}-2A_{ln}^{\left(t\right)}A_{kn}^{\left(t\right)}C_{kn}C_{ln}}{A_{kn}^{\left(t\right)}A_{ln}^{\left(t\right)}}\nonumber \\
 & =\frac{1}{2}\sum_{k=1}^{K}\sum_{n=1}^{N}\sum_{l=1}^{K}\left(\mtbldM^{T}\mtbldM\right)_{kl}\frac{\left(A_{ln}^{\left(t\right)}C_{kn}-A_{kn}^{\left(t\right)}C_{ln}\right)^{2}}{A_{kn}^{\left(t\right)}A_{ln}^{\left(t\right)}}\geq0\label{eq:auxProofCha_1}
\end{align}
Then we compare the second term in Eq.~\eqref{eq:compareAuxAndTaylor_v2}
\begin{equation}
f_{2}=\sum_{k=1}^{K}\sum_{n=1}^{N}\left(\lambda D_{nn}+\frac{\alpha}{A_{kn}^{\left(t\right)}}\right)C_{kn}^{2}-\lambda\text{Tr}\left(\mtbfC\mtbldL\mtbfC^{T}\right).\label{eq:compareAuxAndTaylor_T2}
\end{equation}
When the equation $\mtbfL=\mtbfD-\mtbfW$ is considered, $\text{Tr}\left(\mtbfC\mtbldL\mtbfC^{T}\right)$
could be rewritten as
\begin{equation}
\text{Tr}\left(\mtbfC\mtbldL\mtbfC^{T}\right)=\sum_{n=1}^{N}\sum_{k=1}^{K}D_{nn}C_{kn}^{2}-\sum_{n=1}^{N}\sum_{m=1}^{N}\sum_{k=1}^{K}C_{kn}C_{km}W_{mn}.\label{eq:auxSecondTerm}
\end{equation}
 Because of the inequality,
\begin{align*}
\sum_{n=1}^{N}\sum_{m=1}^{N}\sum_{k=1}^{K}C_{kn}C_{km}W_{mn} & \geq\frac{1}{2}\sum_{n=1}^{N}\sum_{m=1}^{N}\sum_{k=1}^{K}W_{mn}\left(-C_{kn}^{2}-C_{km}^{2}\right)\\
 & =-\sum_{n=1}^{N}\sum_{k=1}^{K}\left(\sum_{m=1}^{N}W_{mn}\right)C_{kn}^{2}=-\sum_{n=1}^{N}\sum_{k=1}^{K}D_{nn}C_{kn}^{2},
\end{align*}
Eq.~\eqref{eq:auxSecondTerm} has the following relationship,
\begin{equation}
\text{Tr}\left(\mtbfC\mtbldL\mtbfC^{T}\right)\leq2\sum_{n=1}^{N}\sum_{k=1}^{K}D_{nn}C_{kn}^{2}.\label{eq:inequation_2}
\end{equation}
Thus Eq.~\eqref{eq:compareAuxAndTaylor_T2} could be relaxed as 
\begin{align}
f_{2} & \geq\sum_{k=1}^{K}\sum_{n=1}^{N}\left(\lambda D_{nn}+\frac{\alpha}{A_{kn}^{\left(t\right)}}\right)C_{kn}^{2}-2\lambda\sum_{n=1}^{N}\sum_{k=1}^{K}D_{nn}C_{kn}^{2}\nonumber \\
 & =\sum_{k=1}^{K}\sum_{n=1}^{N}\left(\frac{\alpha}{A_{kn}^{\left(t\right)}}-\lambda D_{nn}\right)C_{kn}^{2}.\label{eq:auxProofCha_2}
\end{align}
For simplicity, we ignore the impact of $C_{kn}^{2}$ and obtain
\begin{eqnarray}
f_{2}' & = & \sum_{k=1}^{K}\sum_{n=1}^{N}\frac{\alpha}{A_{kn}^{\left(t\right)}}-\sum_{k=1}^{K}\sum_{n=1}^{N}\lambda D_{nn}\geq\frac{\alpha}{u}-KN\lambda v.\label{eq:auxProofCha_3}
\end{eqnarray}
where $u=\min\left\{ A_{kn}^{\left(t\right)}\right\} $ and $v=\max\left\{ D_{nn}\right\} $
for $n=\left\{ 1,\cdots,N\right\} ,\ k=\left\{ 1,\cdots,K\right\} $.
Because of the sparse distribution of the column vectors in $\mtbfA$
and the sparse constraint on $\mtbfA$, there are always some vector
$\left[A_{1n}^{\left(t\right)},\cdots,A_{Kn}^{\left(t\right)}\right]^{T}$
very sparse, which means $u\rightarrow0$ and $\frac{\alpha}{u}\rightarrow\infty$.
Besides, the values of $\alpha$ and $KN\lambda v$ are positive and
finite, we have $\left(\frac{\alpha}{u}-KN\lambda v\right)\gg0$.
Mostly, it is straightforward that $f_{2}\geq0$ in \eqref{eq:auxProofCha_2}.
Thus, we prove the inequality \eqref{eq:compareAuxAndTaylor} by proving
$f_{1}\geq0$ in \eqref{eq:auxProofCha_1} and $f_{2}\geq0$ in \eqref{eq:auxProofCha_2}.
$G\left(\mtbfA,\mtbfA^{\left(t\right)}\right)$ is an auxiliary function
(upper bound function) for $F\left(\mtbfA\right)$. 
\end{IEEEproof}

\subsection{Comparison with Gradient Descent Method}

The Gradient Descent Method (GDM) \cite{Kivinen_97_JIC_GD} is a widely
used optimization algorithm to find a local minimum for an objective
function. In this subsection, we try to find the relationship between
the updating rules defined in Eqs.~\eqref{eq:updatingRule_M},~\eqref{eq:updatingRule_A}
and that of Gradient Descent Method. For the problem defined in \eqref{eq:objFunctionOfOurAlg},
the basic updating rules for GDM are given by
\begin{equation}
M_{lk}\leftarrow M_{lk}+u_{lk}\frac{\partial\mathcal{O}}{\partial M_{lk}},\qquad A_{kn}\leftarrow A_{kn}+v_{kn}\frac{\partial\mathcal{O}}{\partial A_{kn}}.\label{eq:GD_method}
\end{equation}
The parameters $[u_{lk}]\inSetsR^{L\times K},\ [v_{kn}]\inSetsR^{K\times N}$
are the learning rates. For our problem, there are two conditions
to ensure physical meaningful local minima obtained by the updating
rules in \eqref{eq:GD_method}. First, the values of $u_{lk}$ and
$v_{kn}$ are relatively small to get local minima. Second, the values
of $u_{lk}$ and $v_{kn}$ could ensure the nonnegative property of
$M_{lk}$ and $A_{kn}$ during each iteration. One kind of choices
to determine the learning rates $u_{lk}$ and $v_{kn}$ are as follows:
\begin{table}[t]
\caption{Parameters used in Computational Complexity Analysis. \label{tab:complexPara}}

\centering{}%
\begin{tabular}{|c|l|}
\hline 
Parameters &
Description\tabularnewline
\hline 
 $L$ &
the number of bands\tabularnewline
\hline 
$K$ &
the number of \emph{endmembers}\tabularnewline
\hline 
$N$ &
the number of pixels in the hyperspectral image\tabularnewline
\hline 
$p$ &
the number of pixels in the local window\tabularnewline
\hline 
$q$ &
the percentage of nearest neighbors in the local window \tabularnewline
\hline 
$t$ &
the number of iterations\tabularnewline
\hline 
\end{tabular}
\end{table}
 
\begin{table}[t]
\begin{centering}
\caption{Computational operation counts for NMF and SS-NMF during each iteration.\label{tab:ComplexTab}}

\par\end{centering}

\centering{}%
\begin{tabular}{|c||c|c|c||c|}
\hline 
 &
addition &
multiplication &
division &
overall\tabularnewline
\hline 
\multirow{2}{*}{NMF} &
$2LNK+\frac{3}{2}K^{2}\left(L+N\right)$ &
$2LNK+\frac{3}{2}K^{2}\left(L+N\right)$ &
\multirow{2}{*}{$K\left(L+N\right)$} &
\multirow{2}{*}{$O\left(LNK\right)$}\tabularnewline
 & $-\frac{3}{2}K\left(N+L\right)-2K^{2}$ &
$+K\left(L+N\right)$ &  & \tabularnewline
\hline 
\multirow{2}{*}{SS-NMF} &
$2LNK-\frac{3}{2}K\left(N+L\right)-2K^{2}+$ &
$2LNK+\frac{3}{2}K^{2}\left(L+N\right)+$ &
\multirow{2}{*}{$K\left(L+N\right)$} &
\multirow{2}{*}{$O\left(LNK\right)$}\tabularnewline
 & $\frac{3}{2}K^{2}\left(L+N\right)+K\left(pq+2\right)N$ &
$K\left(L+N\right)+K\left(pq+1\right)N$ &  & \tabularnewline
\hline 
\end{tabular}
\end{table}
 
\begin{itemize}
\item Let $u_{lk}=-M_{lk}/\left(\mtbfM\mtbfA\mtbfA^{T}\right)_{lk}$, we
have
\[
M_{lk}+u_{lk}\frac{\partial\mathcal{O}}{\partial M_{lk}}=M_{lk}\frac{\left(\mtbldY\mtbldA^{T}\right)_{lk}}{\left(\mtbldM\mtbldA\mtbldA^{T}\right)_{lk}}.
\]

\item Let $v_{kn}=-A_{kn}/\left(\mtbfM^{T}\mtbfM\mtbfA+2\lambda\mtbfA\mtbfD+\alpha\right)$,
we have
\end{itemize}
\begin{align*}
A_{kn}+v_{kn}\frac{\partial\mathcal{O}}{\partial A_{kn}} & =A_{kn}\frac{\left(\mtbldM^{T}\mtbldY+2\lambda\mtbldA\mtbldW\right)_{kn}}{\left(\mtbldM^{T}\mtbldM\mtbldA+2\lambda\mtbldA\mtbldD+\alpha\right)_{kn}}.
\end{align*}

It is straightforward that Eqs.~\eqref{eq:updatingRule_M} and~\eqref{eq:updatingRule_A}
are special updating rules that could be obtained by the Gradient
Descent Method.

\subsection{Computational Complexity Analysis}

In this subsection, we compare the computational complexity of the
proposed SS-NMF with that of the NMF algorithm. Since the SS-NMF and
NMF algorithms are solved in an iteration way. We describe the complexity
analysis in two steps. First, we analyze the computational complexity
for each iteration; Second, the number of iteration steps are considered.

Besides the updating rules, SS-NMF needs about $O\left(LNp\right)$
to construct the structure relationships between pixel pairs. Suppose
that totally $t$ iterations are needed to get a convergence result.
The total computational cost for NMF is about 
\begin{equation}
O\left(tLNK\right).
\end{equation}
 The total computational cost for SS-NMF is about 
\begin{equation}
O\left(tLNK+LNp\right).
\end{equation}
As can be seen, the computational complexity of SS-NMF is a bit more
than that of NMF when both methods need $t$ iterations. However,
we will show that the iterations needed by SS-NMF are less than that
of NMF in subsection~\ref{sub:Convergence-Study}. In conclusion,
the total computational complexity for SS-NMF is very close to that
of NMF. 

Usually, the big $O$ notation \cite{Thomas_09_MIT_AlgorithmsIntroduction}
is used to analyze the complexity of an algorithm. To be clear, the
precise arithmetic operations as well as the complexity analysis in
the big $O$ notation for each iteration are summarized in Table~\ref{tab:ComplexTab}.
Table~\ref{tab:complexPara} lists the parameters used in the complexity
analysis. There are three kinds of arithmetic operations in each iteration:
addition, multiplication and division. All of them are calculated
in the float-point precision. Specifically, two tips are essential
to get the results in Table~\ref{tab:ComplexTab}. For tip 1, the
order of the matrix multiplication is important. Taking the matrix
multiplication $\mtbldM\mtbldA\mtbldA^{T}$ for example, there are
two orders: $\left(\mtbldM\mtbldA\right)\mtbldA^{T}$ and $\mtbldM\left(\mtbldA\mtbldA^{T}\right)$.
The former approximately needs $2LNK$ AM (Addition and Multiplication)
and the latter costs $\left(N+L\right)K^{2}$ AM (Addition and Multiplication).
Since $K\ll\text{min}\left\{ N,L\right\} $, we have $2LNK\gg\left(N+L\right)K^{2}$,
which means that $\left(\mtbldM\mtbldA\right)\mtbldA^{T}$ is much
more complex than $\mtbldM\left(\mtbldA\mtbldA^{T}\right)$. For tip
2, there are symmetry matrices in the computation. If the fact that
$\left(\mtbldA\mtbldA^{T}\right)$ is a symmetry matrix is considered,
the multiplication of $\mtbfM$ with $\left(\mtbldA\mtbldA^{T}\right)$
costs about $\left(\frac{1}{2}L+N\right)K^{2}$ AM (Addition and Multiplication).

\section{Experiments}

In this section, we investigate the effectiveness of the proposed
SS-NMF algorithm for the Hyperspectral Unmixing (HU) problem. Several
experiments are carried out to show that our algorithm is well suited
for the HU problem.

\subsection{Evaluation Metrics}

In order to evaluate the proposed method, we adopt two performance
metrics: the Spectral Angle Distance (SAD) and the Root Mean Square
Error (RMSE), which are widely used by \cite{Qian_11_TGRS_NMF+l1/2,Jia_09_TGRS_ConstainedNMF,LiuXueSong_2011_TGRS_ConstrainedNMF,Jose_05_TGRS_Vca}.
The SAD is used to evaluate the performance of estimated \emph{endmembers},
which is an angle distance between an estimated \emph{endmember} and
its corresponding ground truth. It is defined as follows 
\begin{equation}
\mbox{SAD}\left(\mtbldm,\hat{\mtbldm}\right)=\arccos\left(\frac{\mtbldm^{T}\hat{\mtbldm}}{\|\mtbldm\|\cdot\|\hat{\mtbldm}\|}\right),\label{eq:sadMetri_experiment}
\end{equation}
 where $\mtbldm$ denotes the ground truth of one \emph{endmember},
$\hat{\mtbldm}$ is the corresponding estimated result. The smaller
SAD corresponds to a better performance. The Root Mean Square Error
(RMSE) is used to evaluate the performance of estimated \emph{abundance}
maps. It is given by 
\begin{equation}
\mbox{RMSE}\left(\mtbldz,\hat{\mtbldz}\right)=\left(\frac{1}{N}\|\mtbldz-\hat{\mtbldz}\|_{2}^{2}\right)^{\frac{1}{2}},\label{eq:rmseMetri_experiment}
\end{equation}
where $N$ is the number of pixels in the hyperspectral image, $\mtbldz$
is the ground truth of one \emph{abundance} map, $\hat{\mtbldz}$
denotes the corresponding estimated result. The smaller RMSE corresponds
to a better performance. In the following subsections, the \emph{abundance}
map (i.e. $\mtbfz$) will be showed in two visible ways: in pseudo
color and in gray scale, as shown in Figs.~\ref{fig:Abundance_UrbanAllSNRS},~\ref{fig:Abundance_UrbanTwoWays},~\ref{fig:Abundance_JasperAllSNRS}
and~\ref{fig:Abundance_JasperTwoWays}.

\subsection{Compared Algorithms}

We compare the proposed method with seven related methods on several
data sets. The details of these evaluated algorithms (including our
algorithm) are listed as follows 
\begin{enumerate}
\item \textbf{Our algorithm}: Structured Sparse regularized Nonnegative
Matrix Factorization (SS-NMF in short) is a new algorithm proposed
in this paper.
\item Vertex Component Analysis (VCA in short) \cite{Jose_05_TGRS_Vca}
is a classic geometrical method that needs the existence of pure pixels
for each \emph{endmember. }Different with the other algorithms that
estimate the \emph{endmembers} and \emph{abundances} simultaneously,
VCA can only estimate the \emph{endmembers. }The \emph{abundances}
are estimated\emph{ }by solving a constrained Least Square Problem
\cite{Heinz_01_ITGRS_FLSQ_HU}. The code for this algorithm is obtained
from ``http://www.lx.it.pt/bioucas/code.htm''. 
\item Nonnegative Matrix Factorization (NMF in short) \cite{Lee_99_Nature_NMF}
is a typical statistical method. Due to the nonnegative constraint,
which could be seen as a special case of sparsity constraint \cite{Qian_11_TGRS_NMF+l1/2},
NMF tends to get parts-based results. The code for this method is
downloaded from ``http://www.ee.columbia.edu/grindlay/code.html''.
\item Nonnegative sparse coding ($\ell_{1}$-NMF in short) is a popular
sparse regularized NMF method proposed by \cite{Hoyer_02_NNSP_NMF_l1}.
The code is available on ``http://www.cs.helsinki.fi/u/phoyer/software.html''. 
\item $\ell_{1/2}$ sparsity-constrained Nonnegative Matrix Factorization
($\ell{}_{\text{1/2}}$-NMF in short) is a state-of-the-art algorithm
for the HU problem. It is proposed by \cite{Qian_11_TGRS_NMF+l1/2}.
Since the code is unavailable on the Internet, we implement it by
ourself.
\item Graph regularized Nonnegative Matrix Factorization (G-NMF in short),
proposed by \cite{Cai_11_PAMI_GNMF}, is a good algorithm to extract
graph information latent in data and transfer it to the low dimension
representation space in the matrix factorization process. The code
is obtained from ``http://www.cad.zju.edu.cn/home/dengcai/Data/GNMF.html''. 
\item Local Neighborhood Weights regularized NMF (W-NMF in short) \cite{JmLiu_12_SlectedTopics_W-NMF}
is a graph based NMF method. The main contribution of this method
is that it integrates the spectral and spatial information when constructing
the weighted graph. Since the code implemented by the author has been
lost, we realize it by ourself. 
\item Endmember Dissimilarity Constrained NMF (EDC-NMF in short) \cite{nWang_13_SelectedTopics_EDC-NMF}
is different from the listed methods. It is the only method that utilizes
a constraint from the endmember prior. With this constraint, it would
be highly possible for EDC-NMF to find a result whose endmember spectra
are smooth and different from each other. The code is implemented
by ourself. 
\end{enumerate}
Among the above eight algorithms, VCA is a geometrical method, the
other seven algorithms belong to statistical ones. There is no parameter
in the NMF and VCA methods. In the next subsection, the methods to
set parameters for $\ell_{1}$-NMF, $\ell_{1/2}$-NMF, G-NMF, W-NMF
and SS-NMF algorithms are introduced. 
\begin{figure}[h]
\begin{centering}
\subfloat[]{\begin{centering}
\includegraphics[scale=0.45]{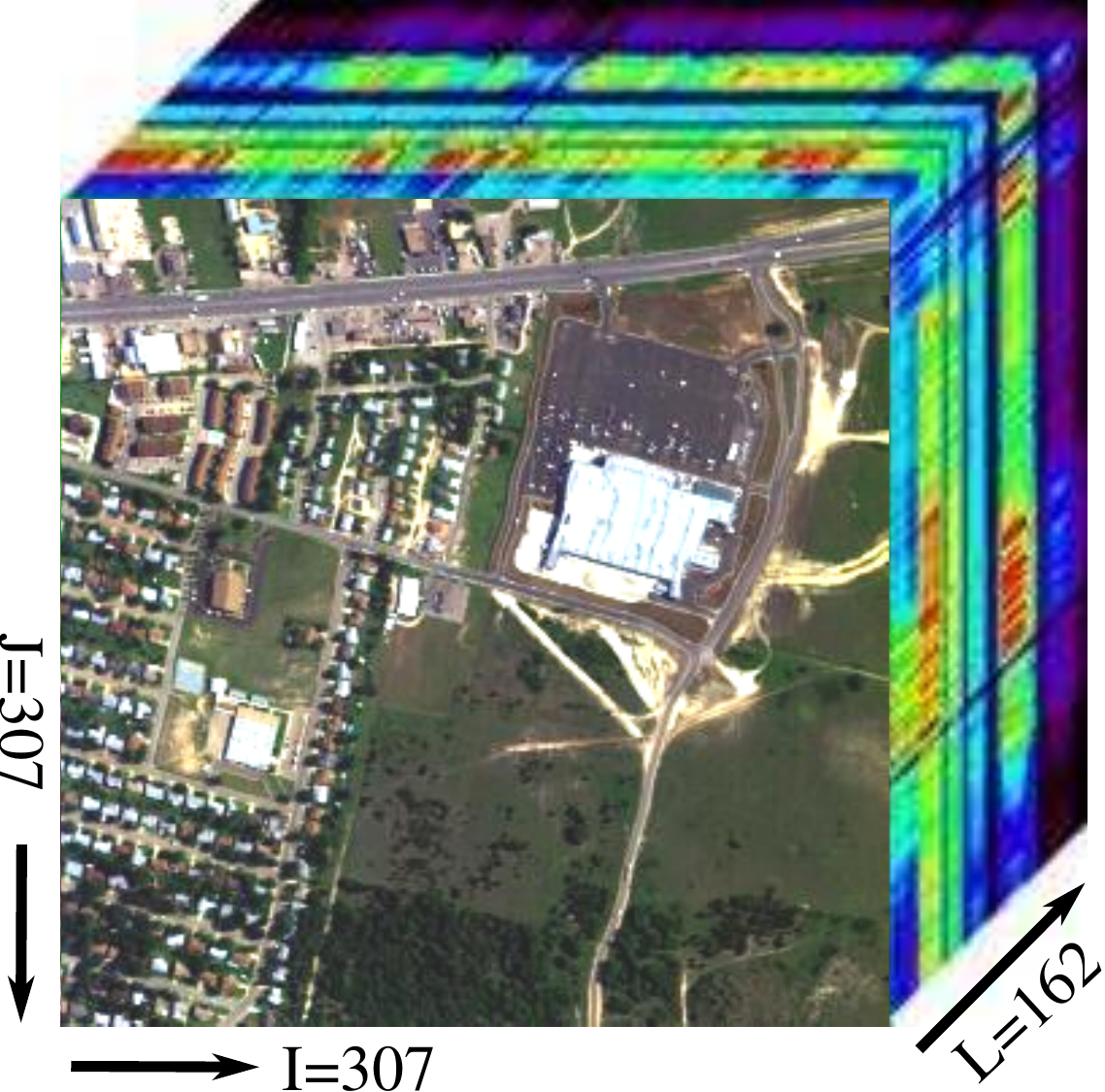}
\par\end{centering}

}\hspace{1cm}\subfloat[]{\centering{}\includegraphics[scale=1.1]{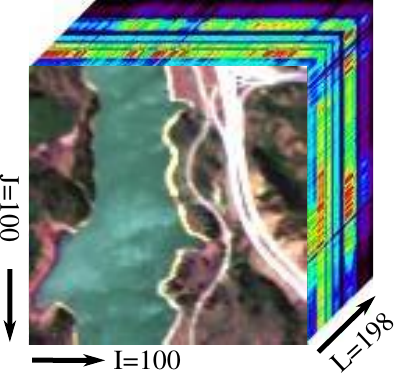}} 
\par\end{centering}

\centering{}\caption{Real hyperspectral data sets used in the experiment: (a) Urban, with
$307\times307$ pixels and $162$ bands; (b) Jasper Ridge, with $100\times100$
pixels and $198$ bands. \label{fig:realHyperImages}}
\end{figure}

\subsection{Parameter Settings }

There are two essential parameters in the proposed SS-NMF method:
$\alpha$ controls the lasso penalty and $\lambda$ controls the structured
constraint. In this subsection, we introduce the methods to set $\alpha$
and $\lambda$ respectively. 

The value of $\alpha$ is closely related to the sparsity of the \emph{abundance
}matrix $\mtbfA$. Unfortunately, $\mtbldA$ is unknown. Here we adopt
a method to estimate $\alpha$ by calculating the sparsity value of
hyperspectral data, which is based on the metric \cite{Hoyer_04_JML_NMFsparse,Qian_11_TGRS_NMF+l1/2}
given by 
\begin{equation}
\alpha_{0}=\frac{1}{\sqrt{L}}\sum_{l=1}^{L}\frac{\sqrt{N}-\|\mathbf{x}_{l}\|_{1}/\|\mathbf{x}_{l}\|_{2}}{\sqrt{N}-1},
\end{equation}
where $\mathbf{x}_{l}\inSetsR_{+}^{N}$ is the image in the $l$-th
channel of hyperspectral data. To improve the accuracy, we set $\alpha$
by searching the range $\left[10^{-1}\alpha_{0},10\alpha_{0}\right]$
at 50 equally spaced values. The parameters in $\ell{}_{\text{1}}$-NMF
and $\ell{}_{\text{1/2}}$-NMF are estimated in the same way. 

One intuitive thought is that the value of $\lambda$ is closely related
to the level of similarities between pixel pairs in the neighborhood.
Thus, the value of $\lambda_{0}$ is estimated in two steps. First,
we randomly select $100$ local patches with $5\times5$ pixels respectively
and compute the similarities between each central pixel and its neighboring
ones. Second, $\lambda_{0}$ is set by averaging all these $100\times(5\times5-1)=2400$
values. To improve the accuracy, we set $\lambda$ by searching the
range $\left[10^{-4}\lambda_{0},10\lambda_{0}\right]$ at 50 equally
spaced values. The parameter to control the strength of the graph
constraint in G-NMF and W-NMF is set similarly.

\subsection{Real Data Sets \label{sub:readDataSets}}

We use two popular hyperspectral data sets \cite{Qian_11_TGRS_NMF+l1/2,Jia_09_TGRS_ConstainedNMF,enviTutorials},
as shown in Fig.~\ref{fig:realHyperImages}, to evaluate the proposed
method. The details of these experiment data sets are introduced in
this subsection. 
\begin{figure}[tb]
\begin{centering}
\includegraphics[width=0.25\columnwidth]{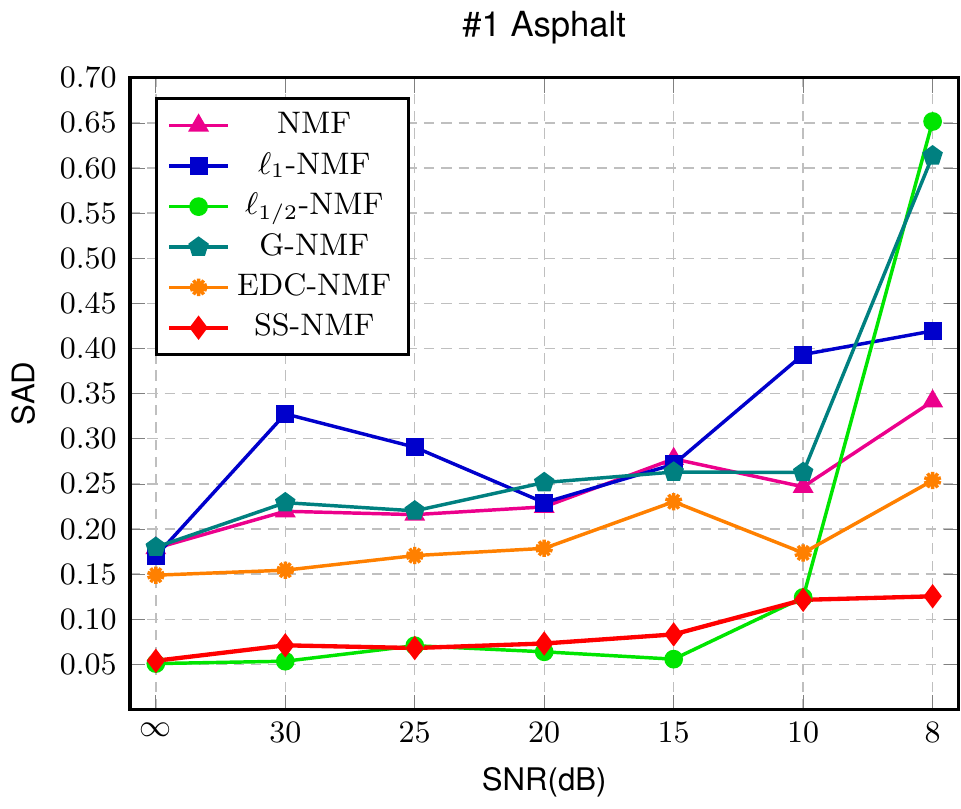}\includegraphics[width=0.25\columnwidth]{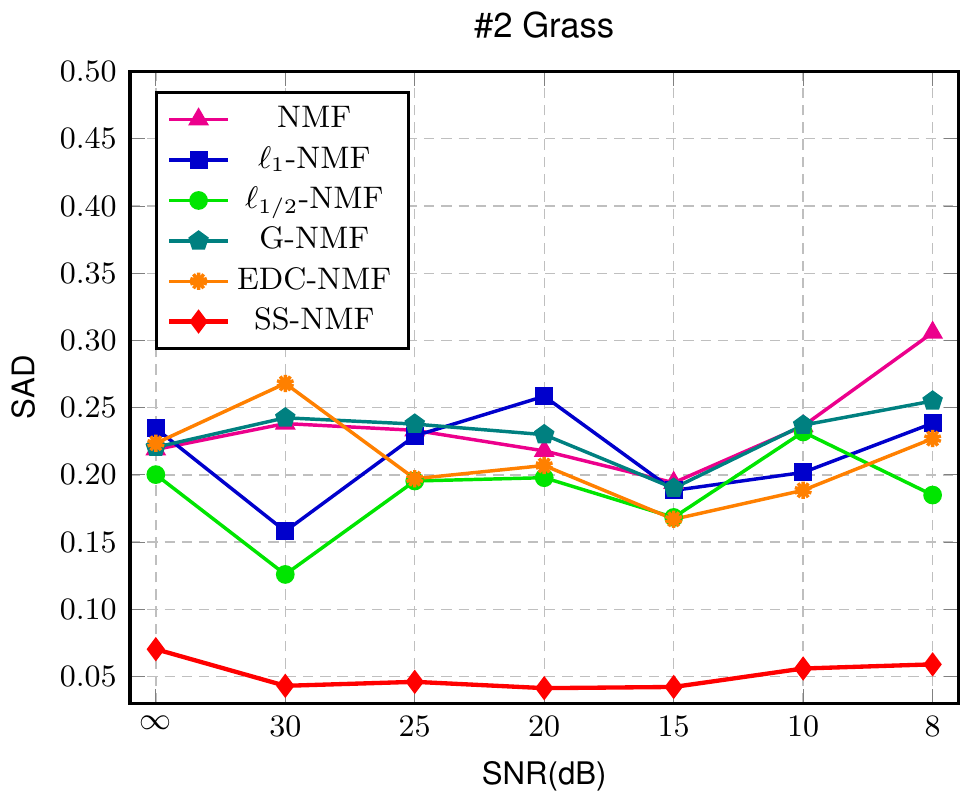}\includegraphics[width=0.25\columnwidth]{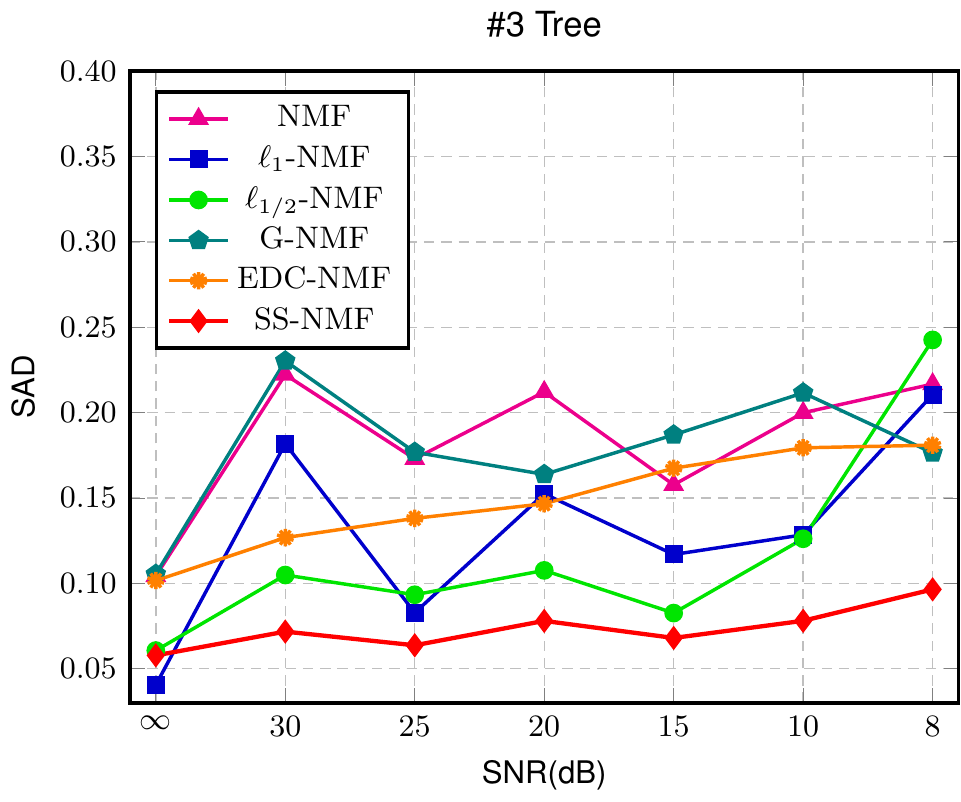}\includegraphics[width=0.25\columnwidth]{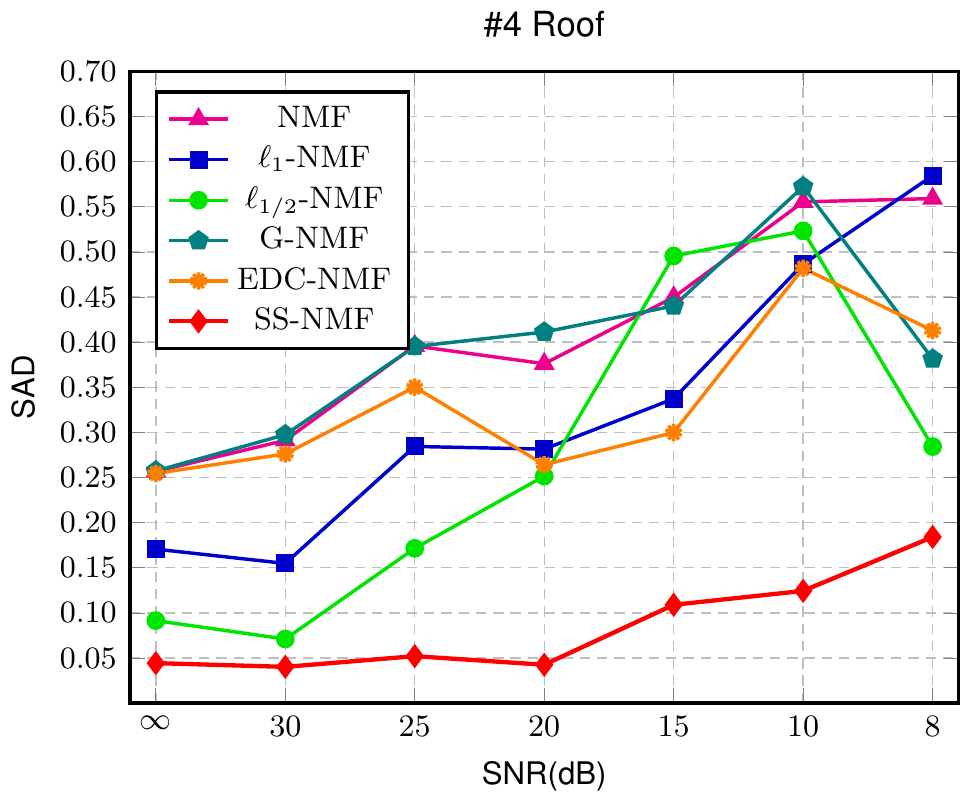}
\par\end{centering}

\centering{}\caption{SADs of four \emph{endmembers} vs. SNRs on the Urban data: (a) Asphalt,
(b) Grass, (c) Tree and (d) Roof. The symbol `$\infty$' at the bottom-left
of each subfigure indicates that no Gaussian noise is added manually.
\label{fig:urban_SAD}}
\end{figure}
 
\begin{table}[tb]
\caption{The average SADs and their standard deviations vs. SNRs on Urban data.
In each row, the value marked in bold is the best. \label{table:urban_mean_SAD} }

\centering{}\includegraphics[width=0.99\columnwidth]{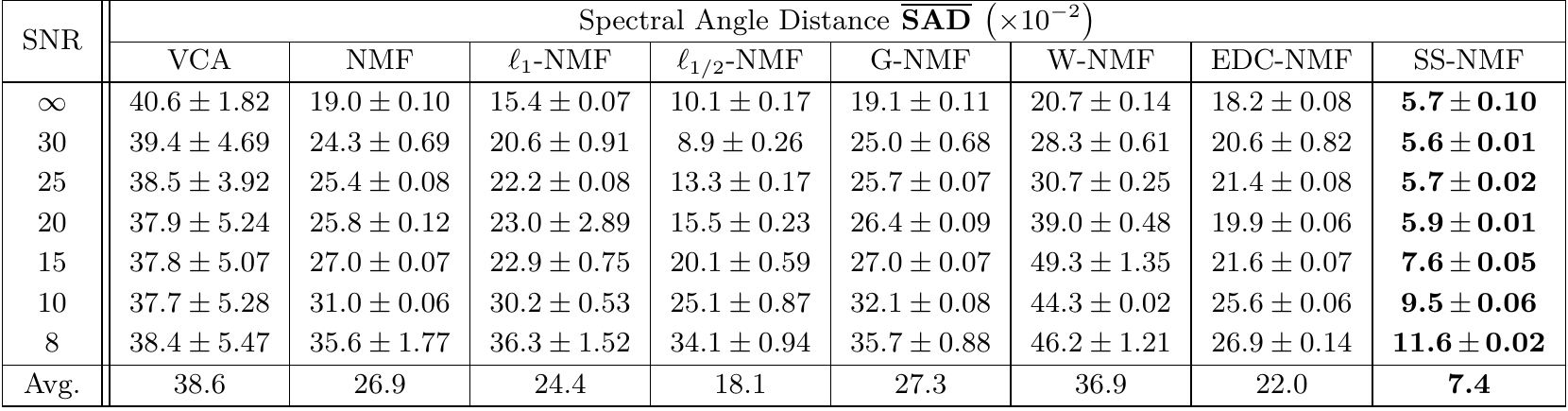}
\end{table}

\textbf{Urban Data}, available at \url{ http://www.tec.army.mil/ Hypercube},
is one of the most widely used hyperspectral data set for the HU research
\cite{Qian_11_TGRS_NMF+l1/2,LiuXueSong_2011_TGRS_ConstrainedNMF}.
It was recorded by the Hyperspectral Digital Imagery Collection Experiment
(HYDICE) in October 1995, whose location is an urban area at Copperas
Cove, TX, U.S. There are $307\times307$ pixels in this image. Each
pixel, corresponding to a $2\times2\ m^{2}$ area, is observed at
$210$ wavelengths ranging from $400$ $nm$ to $2500$ $nm$, with
a spectral resolution of $10$ $nm$. After the bands $1$--$4$,
$76$, $87$, $101$--$111$, $136$--$153$ and $198$--$210$ are
removed (due to dense water vapor and atmospheric effects), $162$
bands are remained in this data. There are four \emph{endmembers}
in this data: \#1 Asphalt, includes the road, the parking area and
a few roofs; \#2 Grass covers areas with a green appearance in Fig.~\ref{fig:realHyperImages}a;
\#3 Tree, owning a different spectral signature with that of grass,
appears dark green mainly in the bottom areas in Fig.~\ref{fig:realHyperImages}a;
\#4 Roof, appears white in the center in Fig.~\ref{fig:realHyperImages}a. 

\textbf{Jasper Ridge }is a popular hyperspectral data set introduced
in \cite{enviTutorials}. There are $512\times614$ pixels in it.
Each pixel is observed at $224$ bands covering the wavelengths ranging
from $380$ $nm$ to $2500$ $nm$. The spectral resolution is $10$
$nm$. As the original hyperspectral data set is too complex to get
the Ground Truth, we consider a subimage with $100\times100$ pixels.
The first pixel corresponds to the pixel $\left(105,269\right)$ in
the original image. After removing bands $1$--$3$, $108$--$112$,
$154$--$166$ and $220$--$224$ (due to dense water vapor and atmospheric
effects), we remain $198$ bands (this is a common preprocess for
the HU analysis). There are four \emph{endmembers} latent in data:
\#1 Road, \#2 Soil, \#3 Water and \#4 Tree, as shown in Fig.~\ref{fig:realHyperImages}b.

The ground truths for both data sets are determined by using the method
proposed by \cite{SJia_07_TGRS_SSCBSS,Jia_09_TGRS_ConstainedNMF}.
There are three steps. First, the VD method, proposed by \cite{Chang_04_TGRS_VDEstNend},
is adopted to determine the endmember number. Second, the endmember
spectra are manually chosen from hyperspectral data. These spectra
have to enjoy high similarity with the reference hyperspectral spectra,
which are supplied by the USGS mineral spectral library %
\footnote{Available on \url{http://speclab.cr.usgs.gov/spectral-lib.html}.%
} etc. Given the \emph{endmembers}, we can get the corresponding \emph{abundances}
by solving a constraint convex optimization problem, which can be
easily implemented by using the Matlab Optimization Toolbox. The ground
truths of the listed data sets are showed in the last column of Figs.~\ref{fig:Abundance_UrbanAllSNRS},
\ref{fig:Abundance_UrbanTwoWays}, \ref{fig:Abundance_JasperAllSNRS}
and \ref{fig:Abundance_JasperTwoWays}. 
\begin{figure}[tb]
\centering{}\includegraphics[width=0.25\columnwidth]{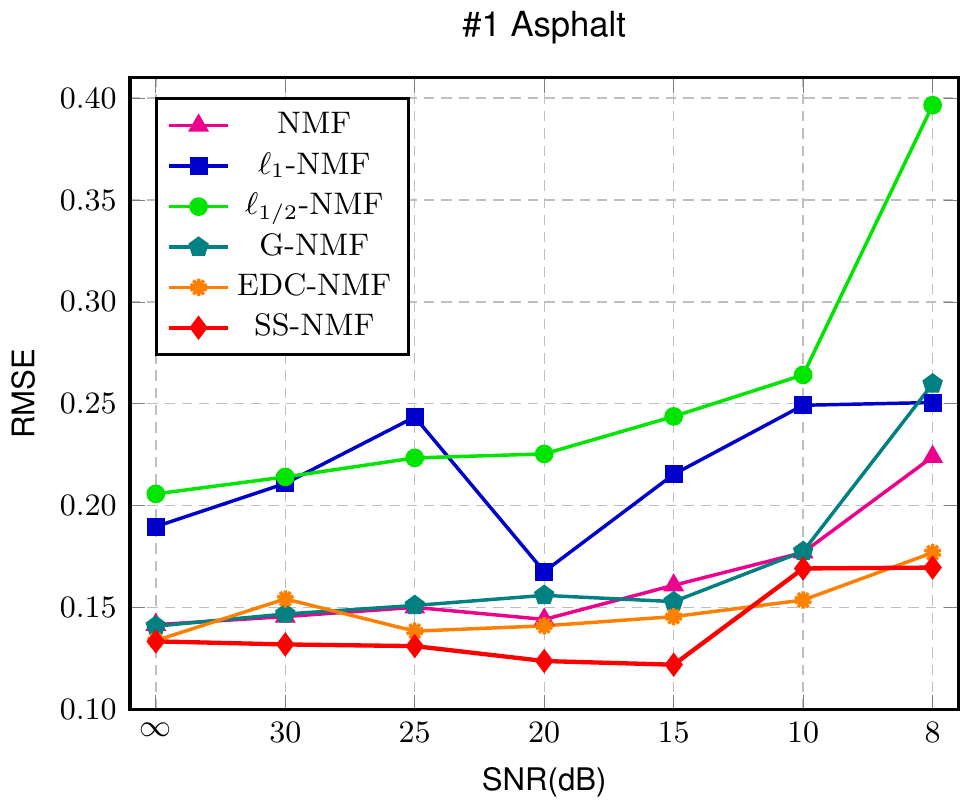}\includegraphics[width=0.25\columnwidth]{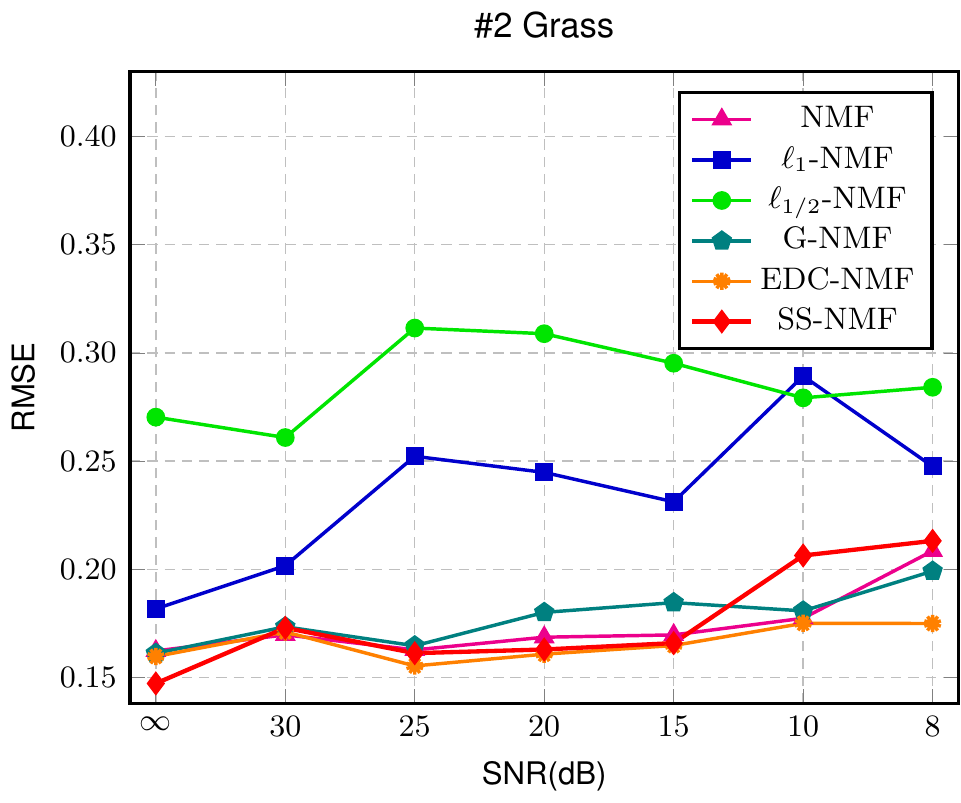}\includegraphics[width=0.25\columnwidth]{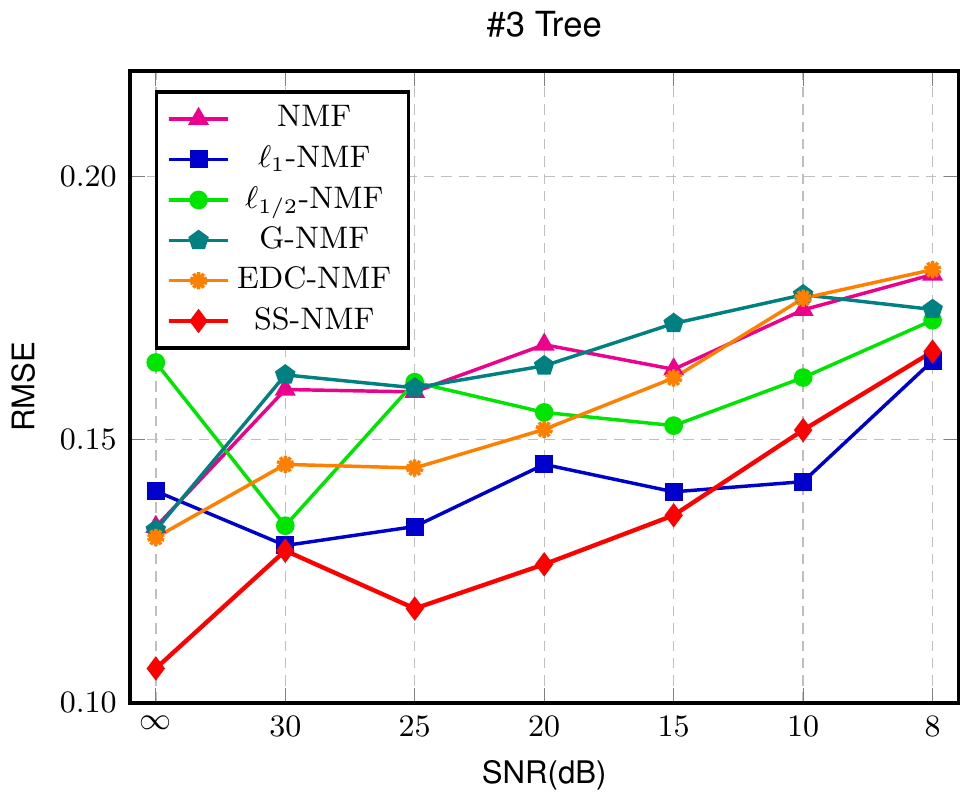}\includegraphics[width=0.25\columnwidth]{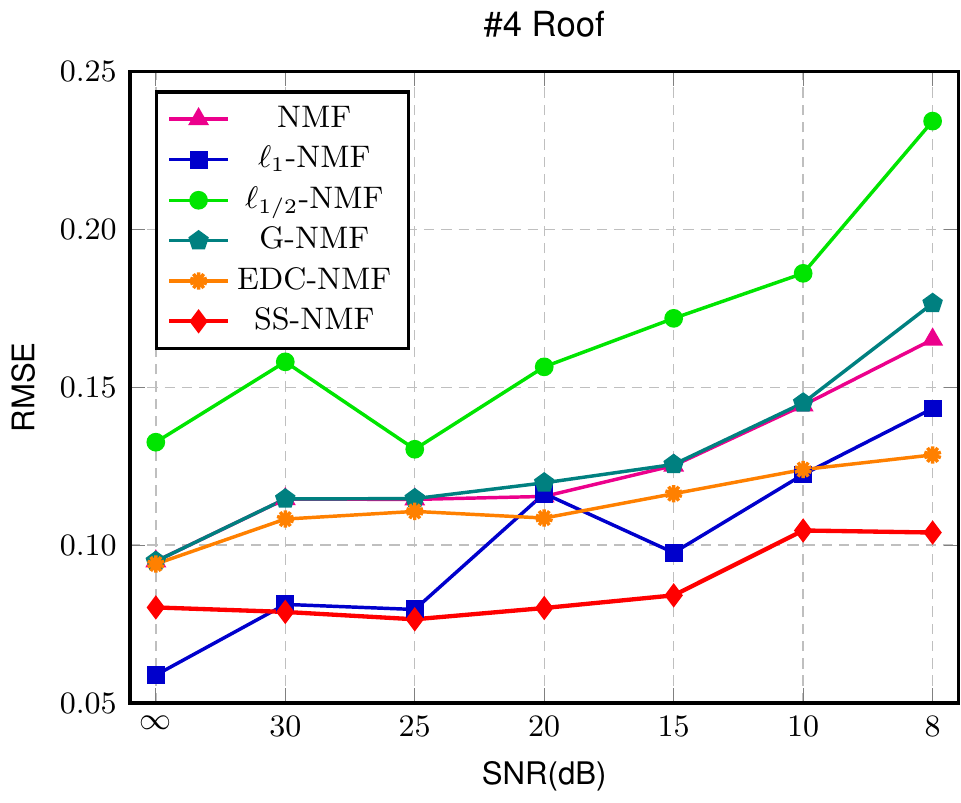}\caption{RMSEs of four \emph{abundance} maps vs. SNRs on the Urban data: (a)
Asphalt, (b) Grass, (c) Tree and (d) Roof. The symbol `$\infty$'
at the bottom-left of each subfigure indicates that no Gaussian noise
is added manually. \label{fig:urban_RMSE}}
\end{figure}
 
\begin{table}[tb]
\caption{The average RMSEs and their standard deviations vs. SNRs on the Urban
data. The values marked in bold are the best. \label{table:urban_mean_RMSE}}

\centering{}\includegraphics[width=0.99\columnwidth]{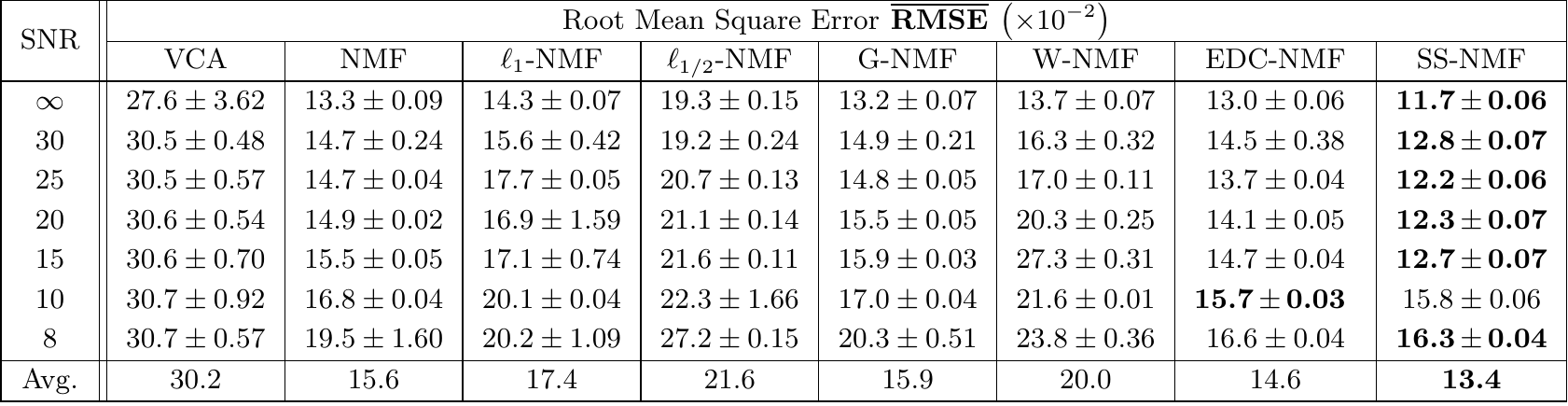}
\end{table}

\subsection{Performance Evaluation}

To test the robustness, this subsection evaluates the influence of
noise on the HU performances for all methods. We choose the i.i.d.
zero-mean white Gaussian noise rather than the correlated noise for
two reasons. First, without any prior, the correlated noise is very
hard to add to the hyperspectral image. Second, the zero-mean white
Gaussian noise is the most widely used noise in the HU study \cite{Qian_11_TGRS_NMF+l1/2,LiuXueSong_2011_TGRS_ConstrainedNMF,nWang_13_SelectedTopics_EDC-NMF}. 

For each hyperspectral data set, seven experiments are carried out
with respect to seven levels of Gaussian noise (i.e., SNR$=\infty$,
$30$, $25$ , $20$, $15$ , $10$, $8$ dB). Each experiment is
repeated 50 times, and the mean results as well as their corresponding
standard deviations are provided here. The evaluation is organized
in two parts: Quantitative Results and Qualitative Results.

\subsubsection{Quantitative Results and Analysis }

Figs.~\ref{fig:urban_SAD},~\ref{fig:urban_RMSE} and Tables~\ref{table:urban_mean_SAD},~\ref{table:urban_mean_RMSE}
show the experiment results on the Urban data set. Figs.~\ref{fig:urban_SAD}
and \ref{fig:urban_RMSE} illustrate the plots of the SAD (metric
for estimated \emph{endmembers}) and the RMSE (metric for estimated
\emph{abundances}) versus seven SNR levels of Gaussian noise respectively.
The results of our proposed SS-NMF algorithm are consistently better
than or comparable to all the other methods. Especially, SS-NMF achieves
significantly better performances in Figs.~\ref{fig:urban_SAD}b,~\ref{fig:urban_SAD}c
and~\ref{fig:urban_SAD}d. 
\begin{figure}[tb]
\noindent \begin{centering}
\includegraphics[width=0.25\columnwidth]{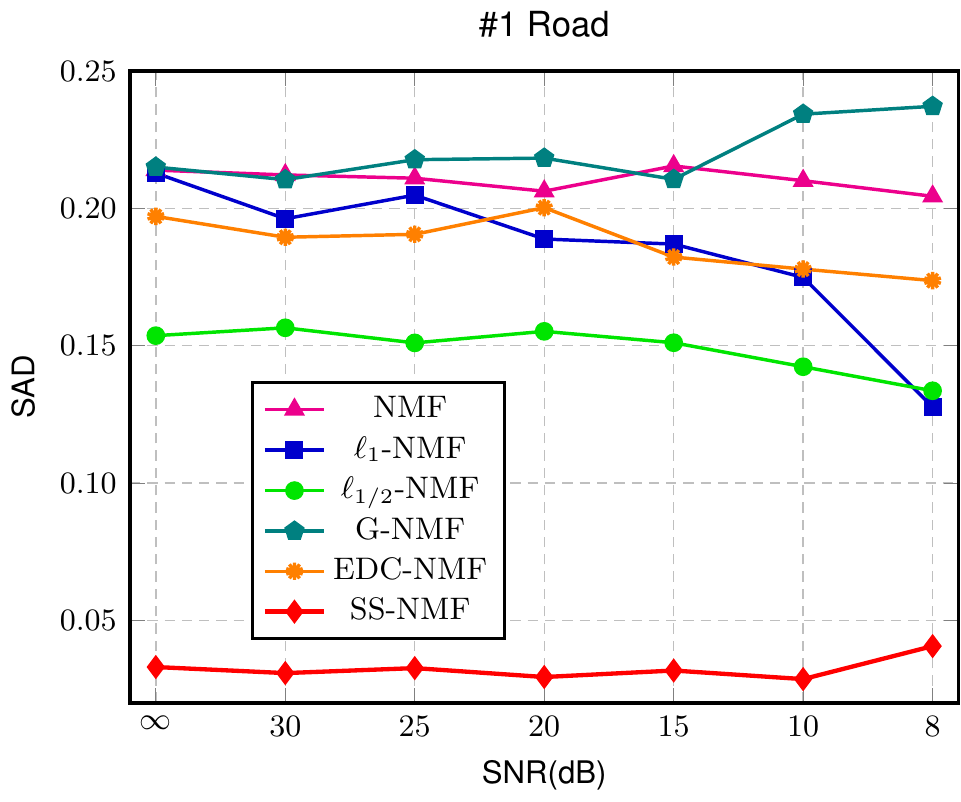}\includegraphics[width=0.25\columnwidth]{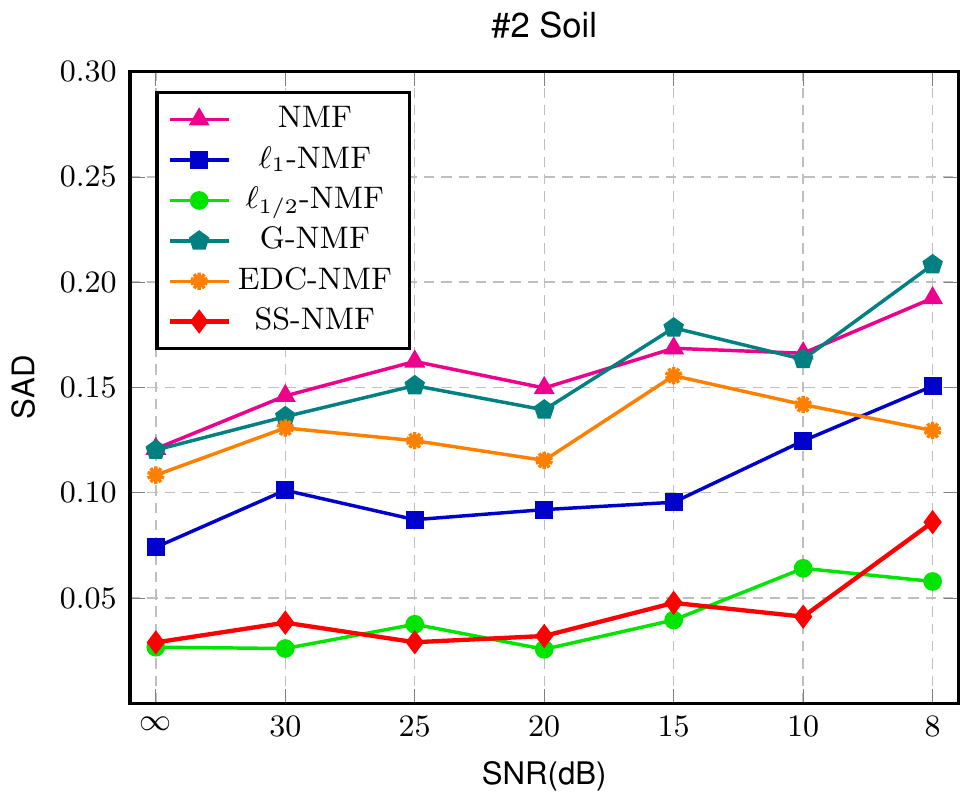}\includegraphics[width=0.25\columnwidth]{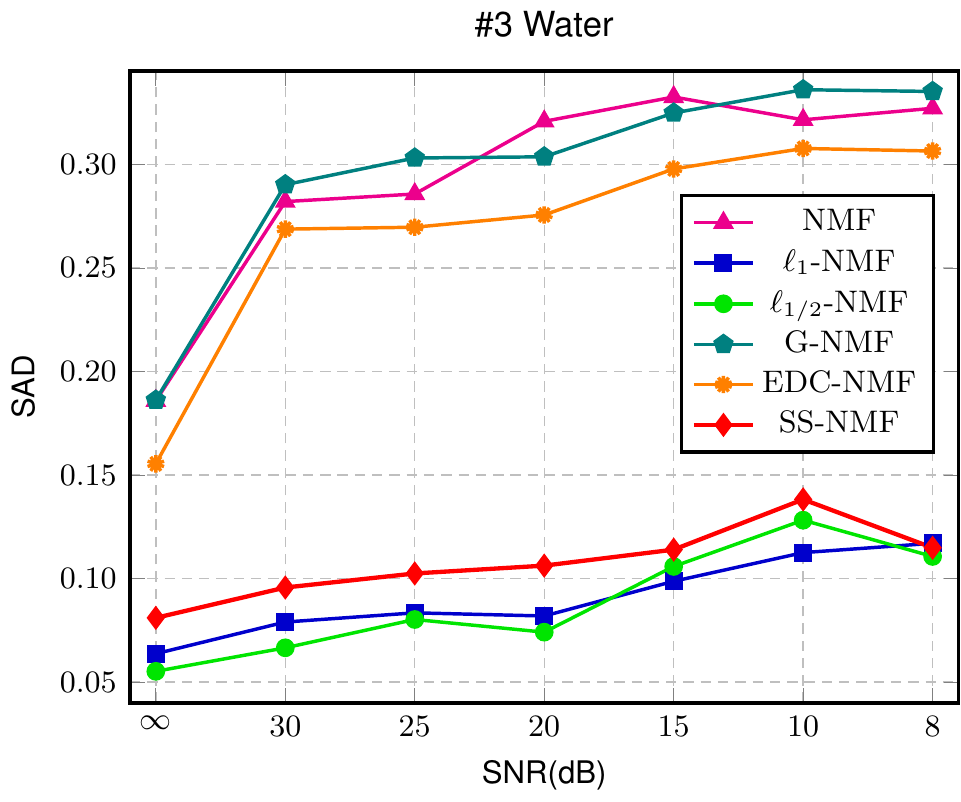}\includegraphics[width=0.25\columnwidth]{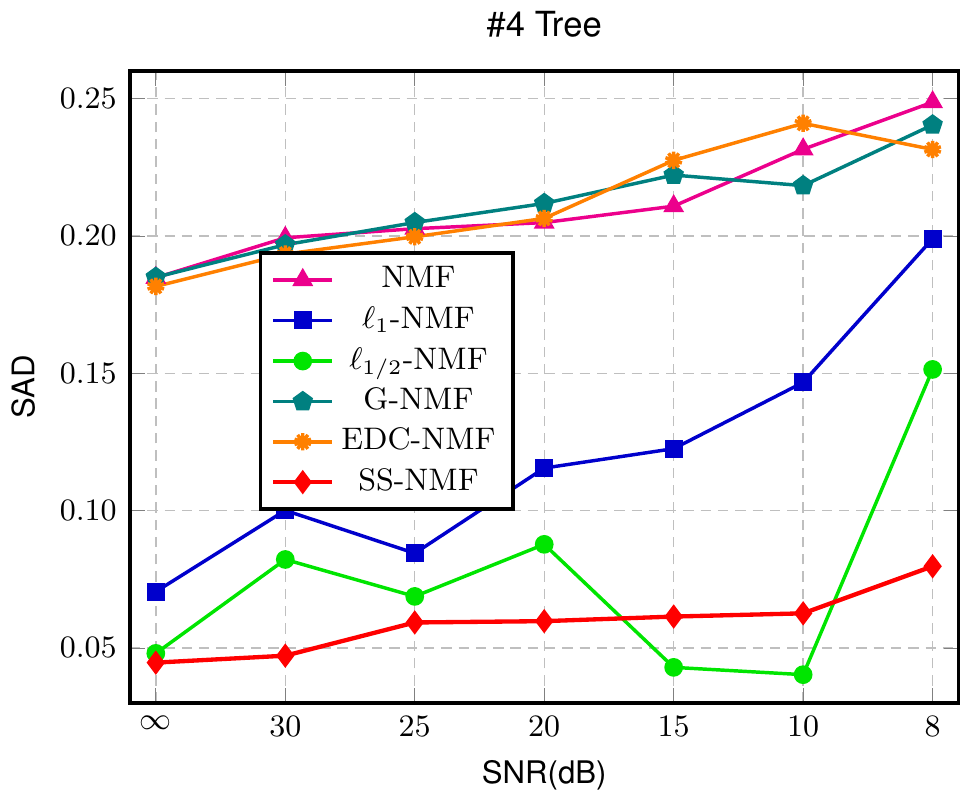}
\par\end{centering}

\noindent \centering{}\caption{SADs of four \emph{endmembers} vs. SNRs on the Jasper Ridge data:
(a) Road, (b) Soil, (c) Water and (d) Tree. The symbol `$\infty$'
at the bottom-left of each subfigure indicates that no Gaussian noise
is added manually. \label{fig:jasper_SAD} }
\end{figure}
\begin{table}[tb]
\caption{The average SADs and their standard deviations vs. SNRs on the Jasper
Ridge data. The values marked in bold are the best. \label{table:jasper_SAD} }

\centering{}\includegraphics[width=0.99\columnwidth]{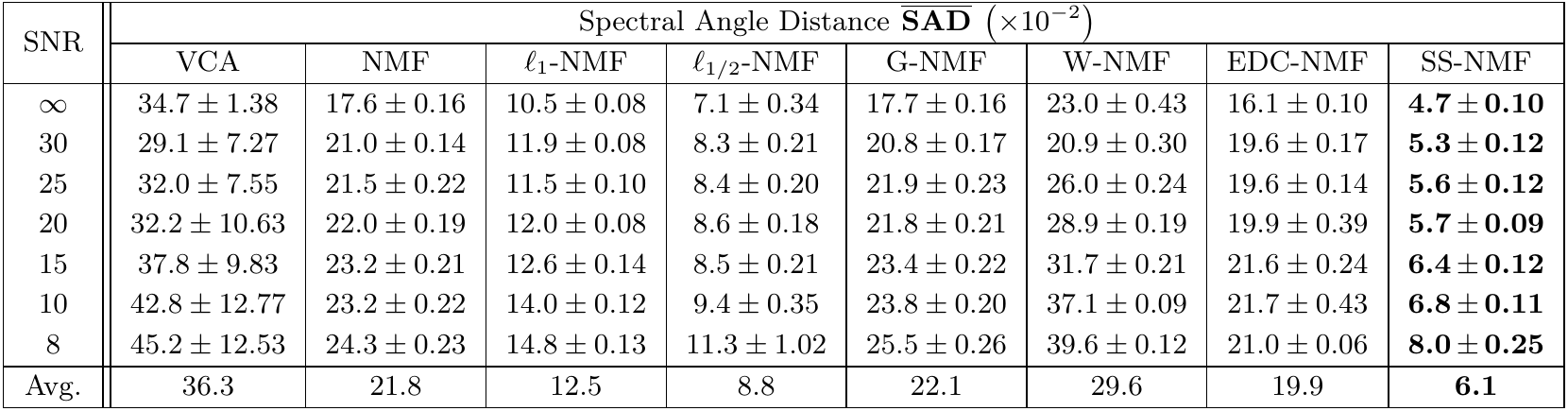}
\end{table}

Tables~\ref{table:urban_mean_SAD} and \ref{table:urban_mean_RMSE}
show the detailed SAD and RMSE performances and their corresponding
standard deviations versus seven levels of Gaussian noise. The values
in Table~\ref{table:urban_mean_SAD} are the average of the statistics
plotted in four subfigures in Fig.~\ref{fig:urban_SAD}. Taking the
value ``\textbf{5.7}'' marked in bold at the northeast corner of Table~\ref{table:urban_mean_SAD}
for example, it is the average of the four SAD values of the SS-NMF
algorithm under the `$\infty$' noise level condition in Fig.~\ref{fig:urban_SAD}.
Similarly, Table~\ref{table:urban_mean_RMSE} illustrates the average
RMSE values in four subfigures in Fig.~\ref{fig:urban_RMSE}. From
the above results, we observe that
\begin{itemize}
\item For all the seven noise levels, our SS-NMF gets the best or comparable
performances. Comparing to the best algorithms other than our proposed
SS-NMF algorithm, i.e. $\ell_{1/2}$-NMF for SAD and EDC-NMF for RMSE
respectively, SS-NMF achieves $59.12\%$ decrement for SAD, as shown
in Table~\ref{table:urban_mean_SAD}, and $8.22\%$ decrement for
RMSE, as shown in Table~\ref{table:urban_mean_RMSE}. Moreover, when
the level of noise varies from SNR=$\infty$ to SNR=$8$, our method
is the most stable one. Those all prove that the structured sparse
regularization is well suited for the HU problem. 
\item Comparing with the results obtained by G-NMF \cite{Cai_11_PAMI_GNMF}
and W-NMF \cite{JmLiu_12_SlectedTopics_W-NMF}, the sparse regularized
NMF methods (i.e., $\ell{}_{\text{1}}$-NMF \cite{Hoyer_02_NNSP_NMF_l1},
$\ell{}_{\text{1/2}}$-NMF \cite{Qian_11_TGRS_NMF+l1/2}) achieve
better or comparable results. This demonstrates that the sparse constraint
is more important than the graph constraint when estimating \emph{endmembers}.
This is mainly because the graph constraint helps to find a new \emph{abundance}
space, in which similar pixels share similar \emph{abundances}. Hence
it leads to smooth or hazy \emph{abundances}, which is closely related
to less expressive \cite{Stanzli_01_CVPR_locNMF} or inaccurate \emph{endmembers}
by the iterative updating rules introduced by \cite{Lee_99_Nature_NMF,Cai_11_PAMI_GNMF}. 
\item NMF and its extensions (SS-NMF and $\ell{}_{\text{1}}$-NMF \cite{Hoyer_02_NNSP_NMF_l1},
$\ell{}_{\text{1/2}}$-NMF \cite{Qian_11_TGRS_NMF+l1/2}, G-NMF \cite{Cai_11_PAMI_GNMF})
outperform the VCA algorithm significantly. There are two reasons
for this: (1) NMF based methods overcome the requirement of pure pixels,
which is essential for the VCA method. (2) The parts-based property
of NMF helps to discover latent \emph{endmembers} and their corresponding
\emph{abundances}. 
\begin{figure}[tb]
\begin{centering}
\includegraphics[width=0.248\columnwidth]{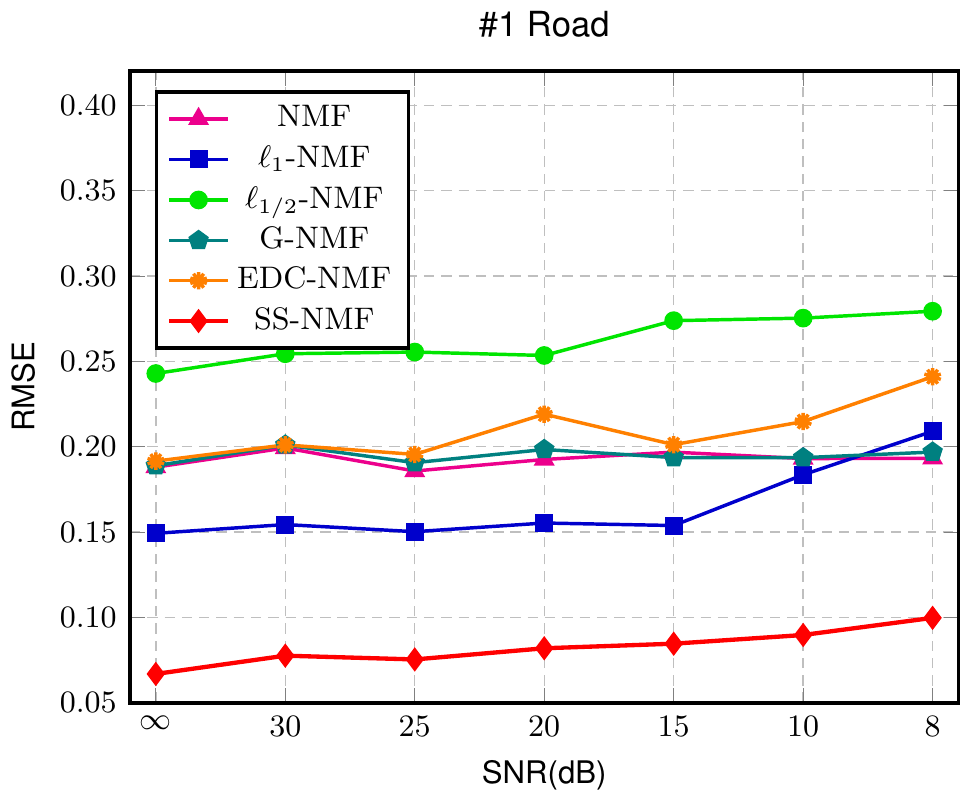}\includegraphics[width=0.248\columnwidth]{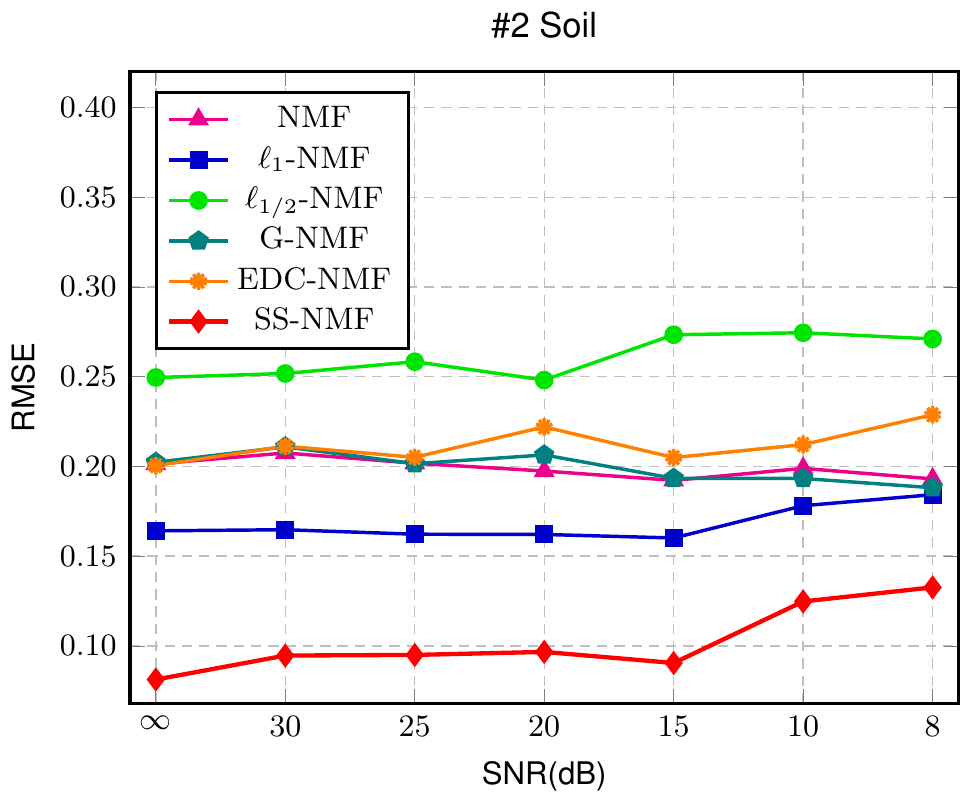}\includegraphics[width=0.248\columnwidth]{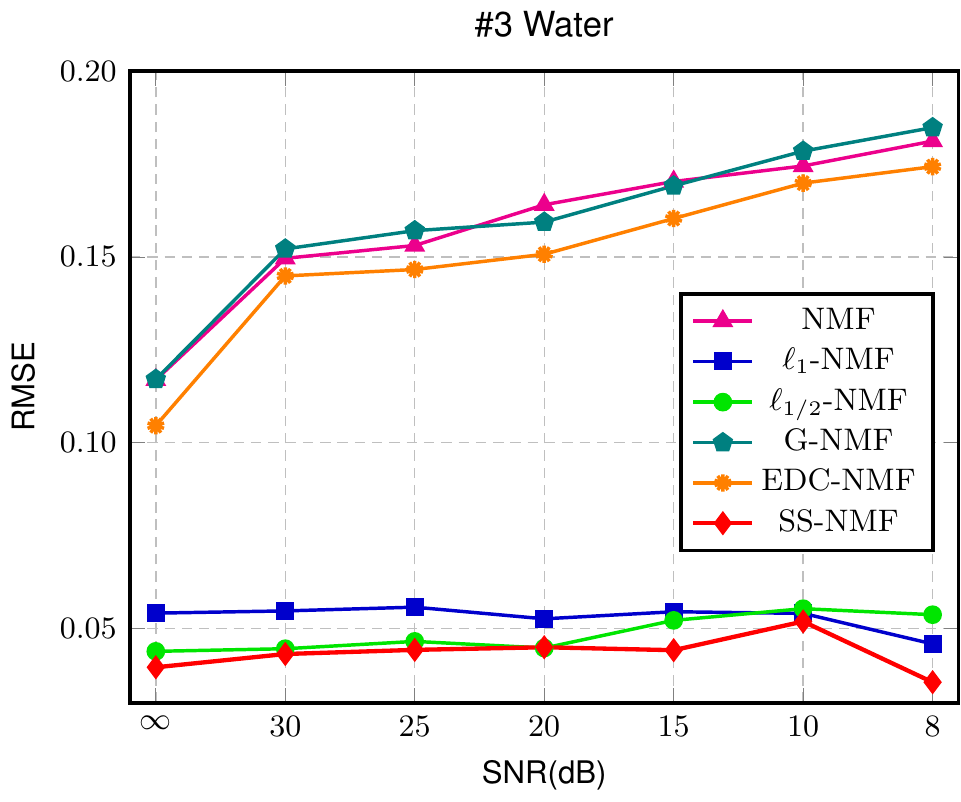}\includegraphics[width=0.248\columnwidth]{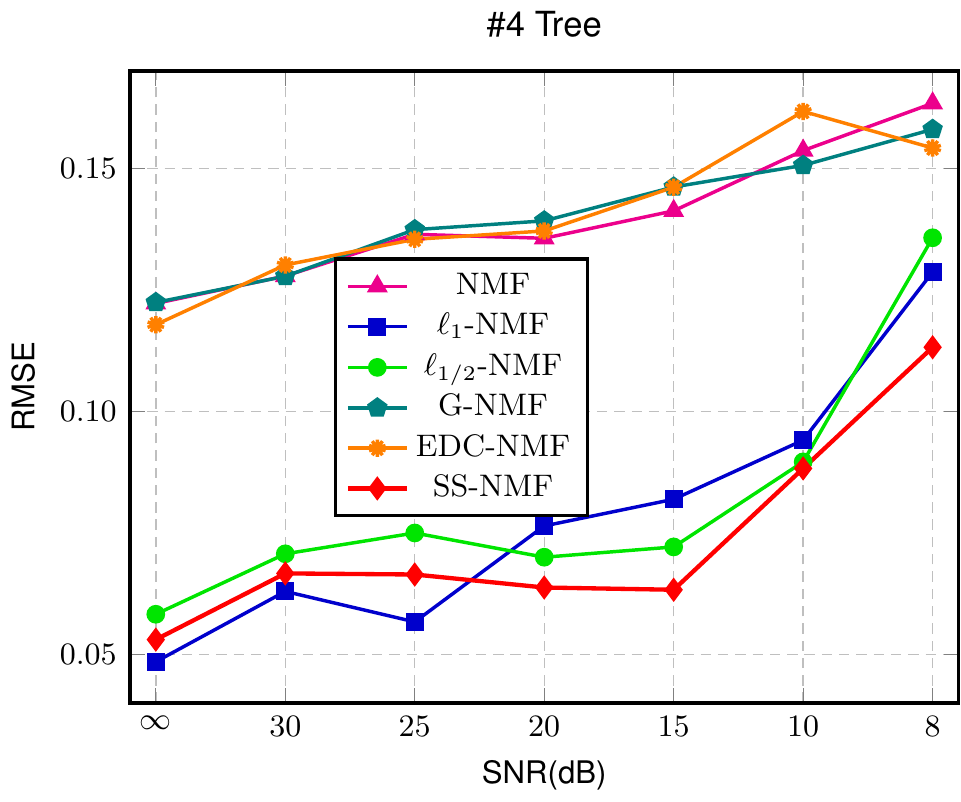}
\par\end{centering}

\centering{}\caption{RMSEs of four \emph{abundance} maps vs. SNRs on the Jasper Ridge data:
(a) Road, (b) Soil, (c) Water and (d) Tree. The symbol `$\infty$'
at the bottom-left of each subfigure indicates that no Gaussian noise
is added manually. \label{fig:jasper_RMSE}}
\end{figure}
\begin{table}[tb]
\caption{The average RMSEs and their standard deviations vs. SNRs on Jasper
Ridge data. The values marked in bold are the best. \label{table:jasper_RMSE} }

\centering{}\includegraphics[width=0.99\columnwidth]{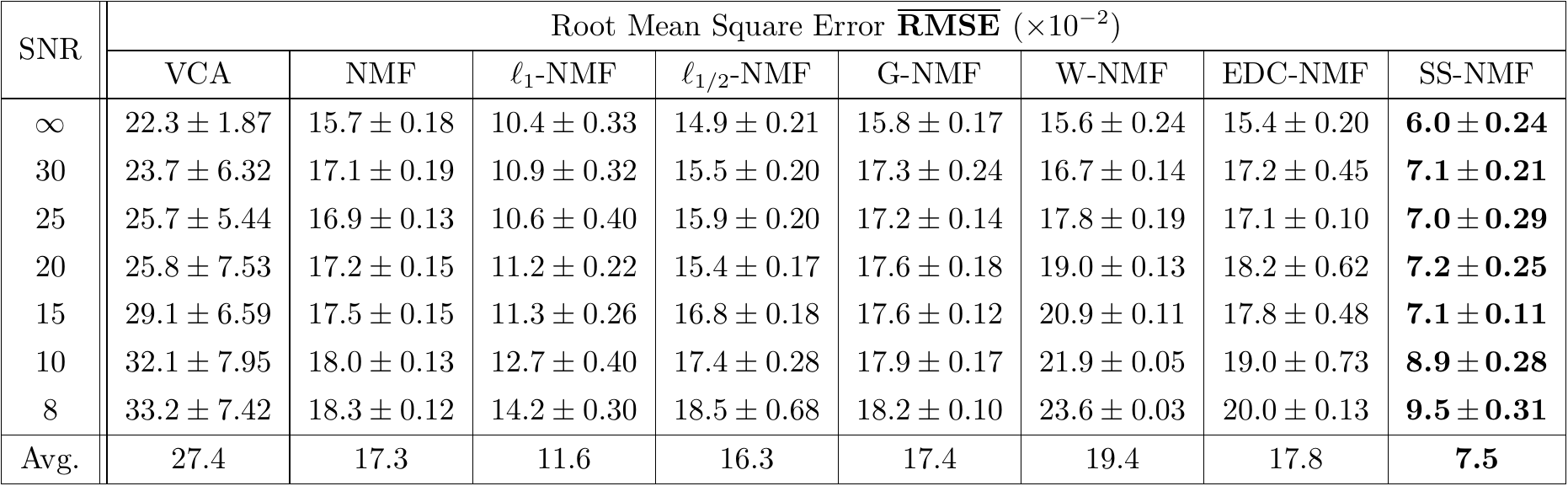}
\end{table}

\end{itemize}
Figs.~\ref{fig:jasper_SAD},~\ref{fig:jasper_RMSE} and Tables~\ref{table:jasper_SAD},~\ref{table:jasper_RMSE}
show the experiment results on the Jasper Ridge data. The HU results
and their standard deviations versus seven levels of noise are summarized
in Tables~\ref{table:jasper_SAD},~\ref{table:jasper_RMSE}. Graphical
plots are shown in Figs.~\ref{fig:jasper_SAD},~\ref{fig:jasper_RMSE}.
As we can see, our proposed SS-NMF are consistently better than or
comparable to all the other methods. In particular, Figs.~\ref{fig:jasper_SAD}a,
\ref{fig:jasper_RMSE}a,~\ref{fig:jasper_RMSE}b and Tables~\ref{table:jasper_SAD},~\ref{table:jasper_RMSE}
show that SS-NMF outperform the other algorithms significantly. Comparing
to the best algorithm other than our proposed algorithm, i.e. $\ell_{1/2}$-NMF
for SAD and $\ell_{1}$-NMF for RMSE respectively, SS-NMF achieves
30.68\% decrement for SAD (metric for \emph{endmember}) in Table~\ref{table:jasper_SAD}
and 35.34\% decrement for RMSE (metric for \emph{abundance}) in Table~\ref{table:jasper_RMSE}.

\subsubsection{Qualitative Results and Analysis}

\begin{figure}[h]
\begin{centering}
\includegraphics[width=0.99\columnwidth]{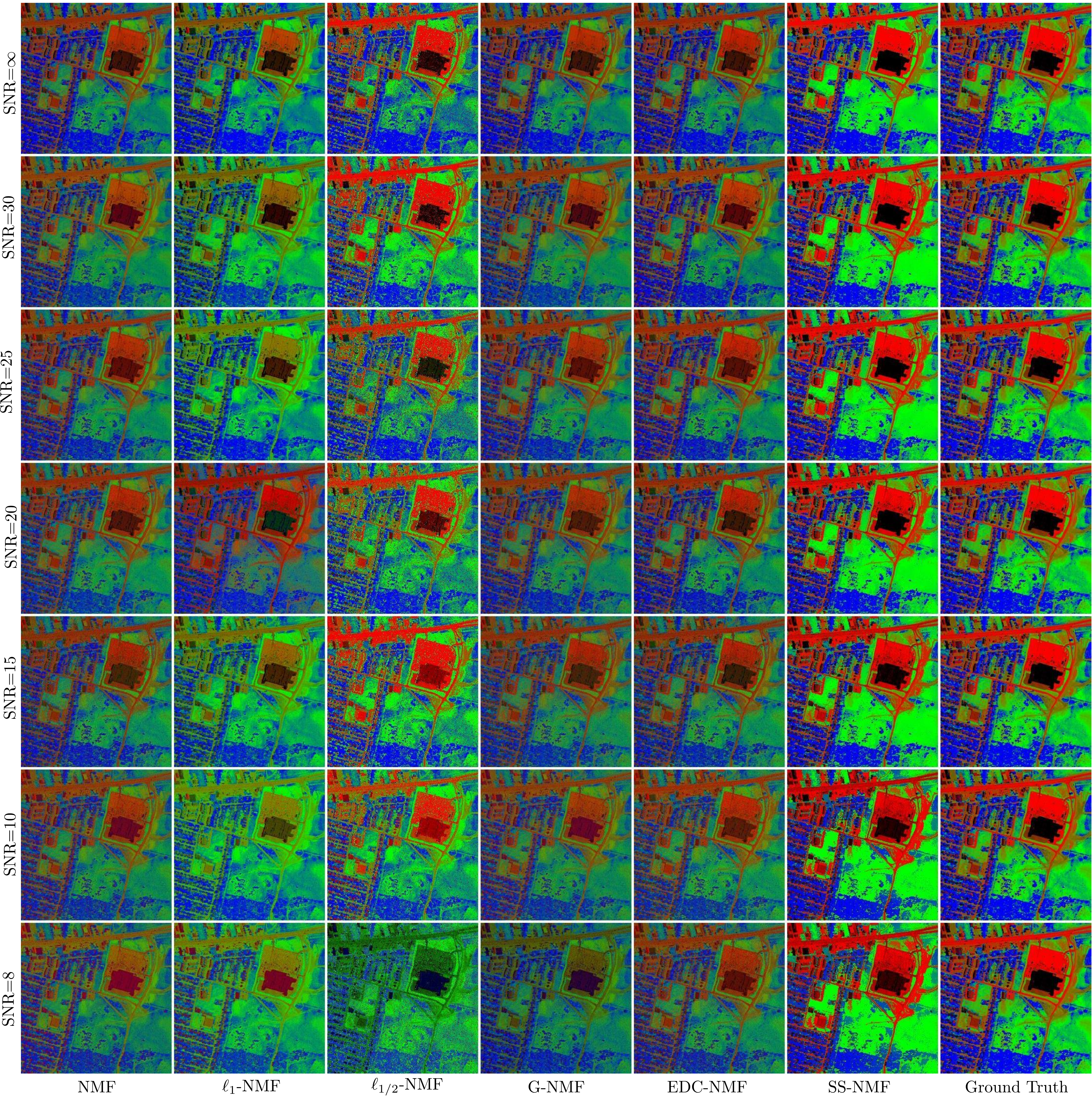}
\par\end{centering}

\caption{\emph{Abundance }maps vs. seven noise levels (SNR$=\infty$, $30$,
$25$ , $20$, $15$ , $10$, $8$ dB) on the Urban data. There are
seven rows and seven columns in this figure. Each row displays the
results under one noise level. From the $1$-st to the $6$-th column,
each column illustrates the results for one algorithm. The last column
shows the Ground Truths. For each subfigure, the proportions of Red,
Blue, Green and Black ink associated with each pixel represent the
fractional \emph{abundances} of Asphalt, Tree, Grass and Roof in the
corresponding pixel. This figure is best viewed in color. \label{fig:Abundance_UrbanAllSNRS} }
\end{figure}

\begin{figure}[h]
\begin{centering}
\subfloat[\emph{Abundance }maps in pseudo color. ]{\centering{}\includegraphics[width=0.99\columnwidth]{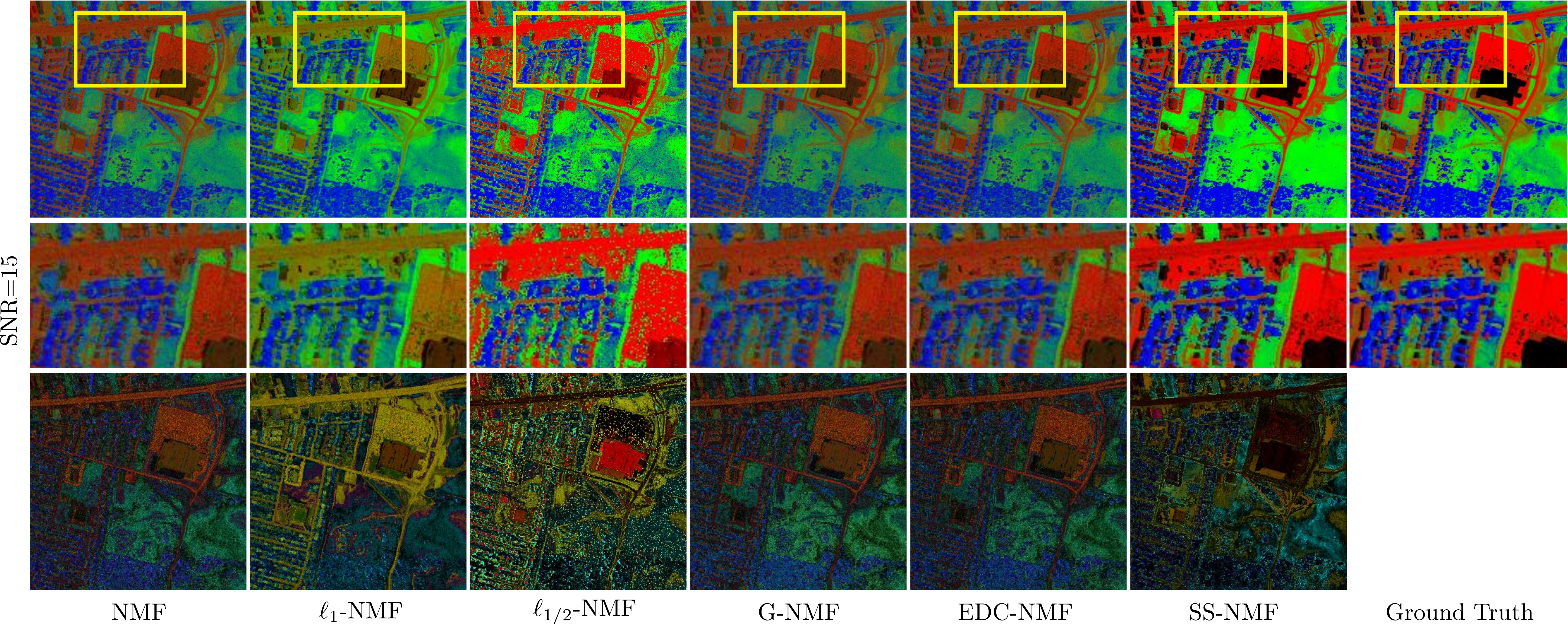}} 
\par\end{centering}

\begin{centering}
\subfloat[\emph{Abundance }maps in gray scale.]{\centering{}\includegraphics[width=0.99\columnwidth]{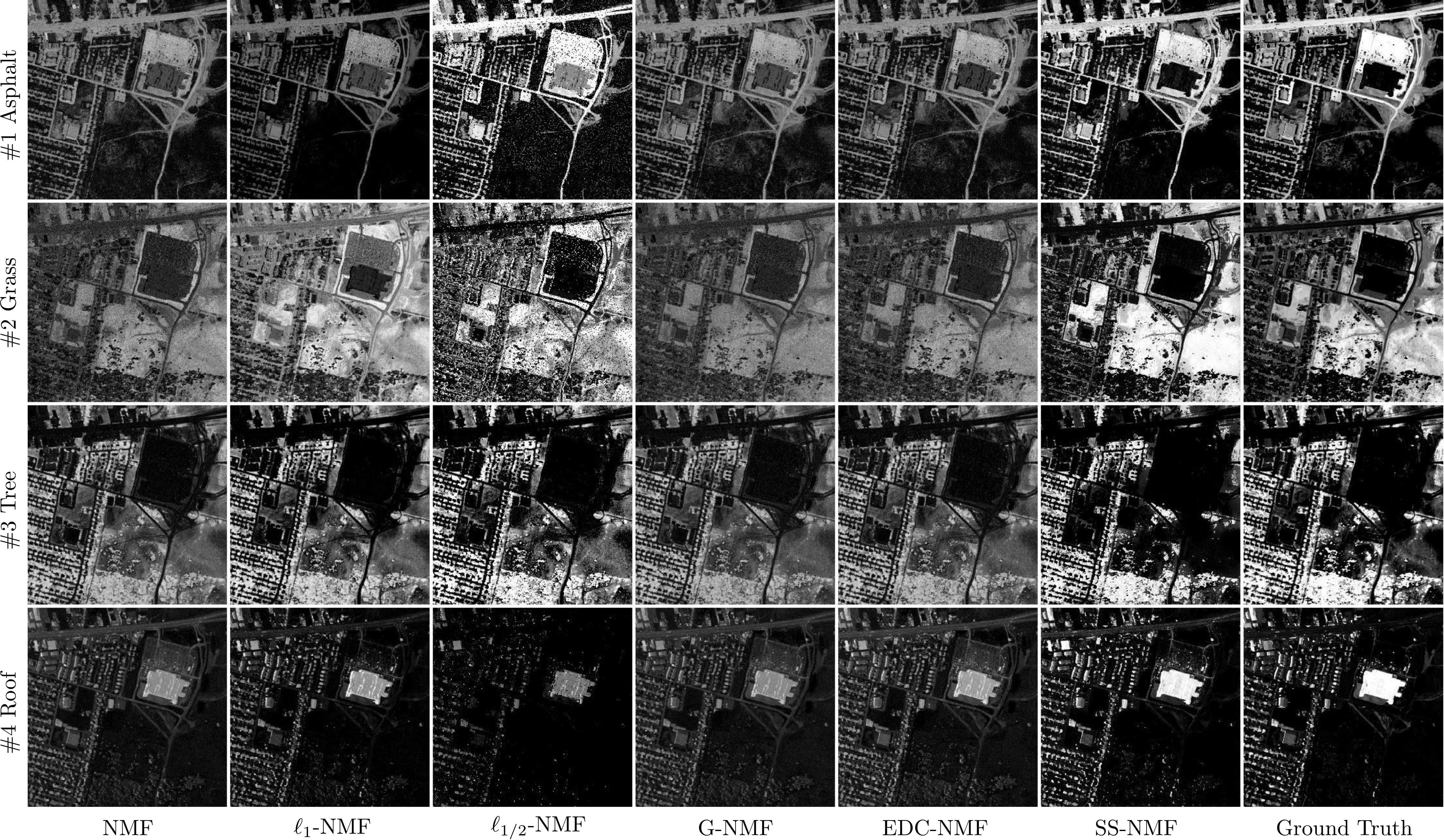}}
\par\end{centering}

\caption{\emph{Abundance }maps under SNR$=15$ dB noise level on the Urban
data: (a) in pseudo color and (b) in gray scale. There are three rows
in (a). The second row displays the results in the yellow box in the
top row. The third row shows the absolute value of the difference
between the estimated results and the ground truth. For each subfigure
in (a), the proportions of Red, Blue, Green and Black ink associated
with each pixel represent the fractional \emph{abundances} of Asphalt,
Tree, Grass and Roof in the corresponding pixels. There are four rows
and seven columns in (b). Each row shows the \emph{abundance }maps
of one target. From the $1$-st to the $6$-th column, each column
illustrates the results of one algorithm. The last column shows the
Ground Truths. This figure is best viewed in color. \label{fig:Abundance_UrbanTwoWays}}
\end{figure}

\begin{figure}[tb]
\begin{centering}
\includegraphics[width=0.99\columnwidth]{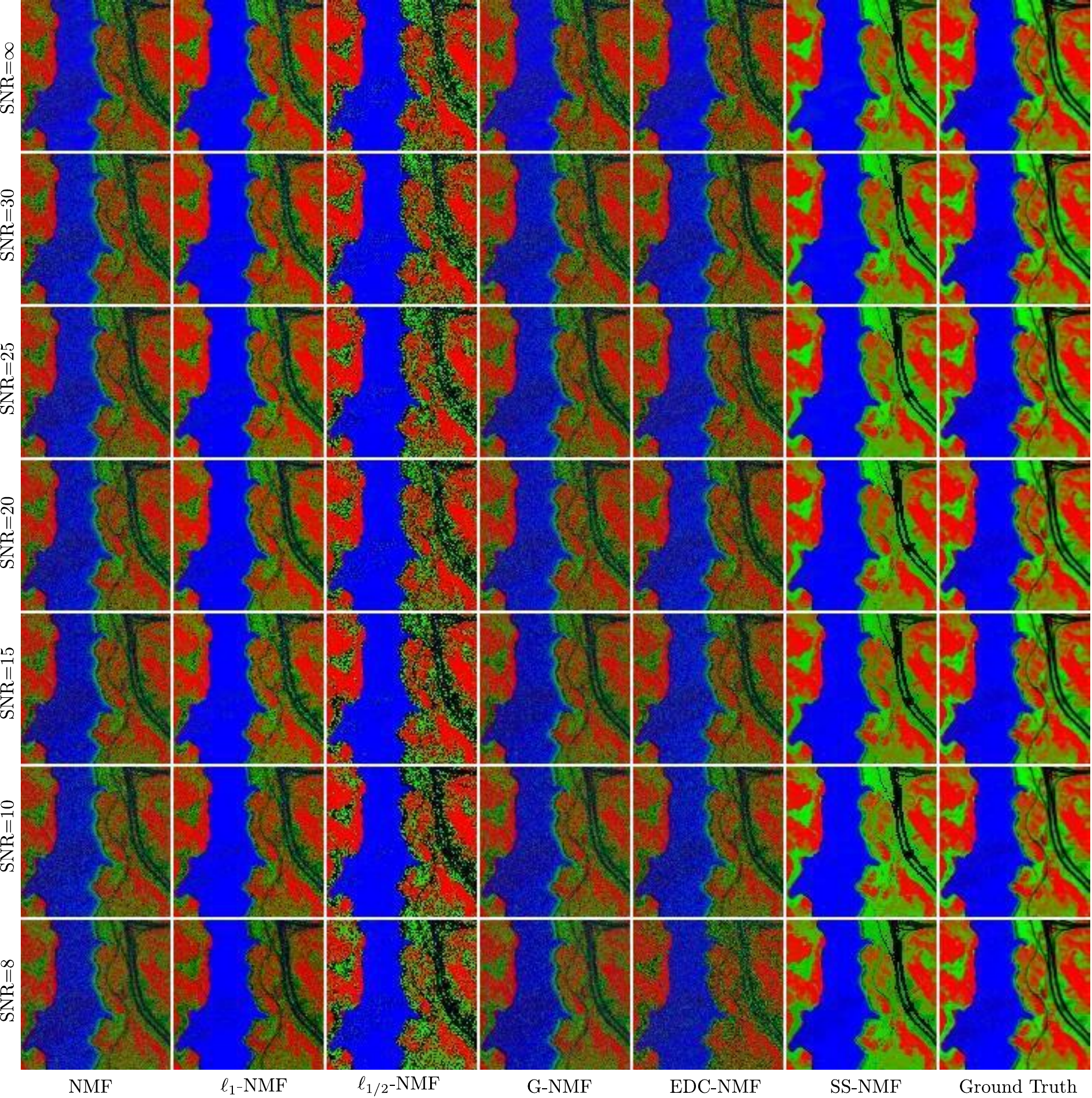}
\par\end{centering}

\caption{\emph{Abundance }maps vs. seven noise levels (SNR$=\infty$, $30$,
$25$ , $20$, $15$ , $10$, $8$ dB) on the Jasper Ridge data. There
are seven rows and seven columns in this figure. Each row displays
the results under one noise level. From the $1$-st to the $6$-th
column, each column shows the results of one method. The last column
illustrates the Ground Truths. For each subfigure, the proportions
of Red, Blue, Green and Black ink associated with each pixel represent
the fractional \emph{abundances} of Tree, Water, Soil and Road in
the corresponding pixel. This figure is best viewed in color. \label{fig:Abundance_JasperAllSNRS}}
\end{figure}

\begin{figure}[tb]
\begin{centering}
\subfloat[\emph{Abundance }maps in pseudo color. ]{\centering{}\includegraphics[width=0.99\columnwidth]{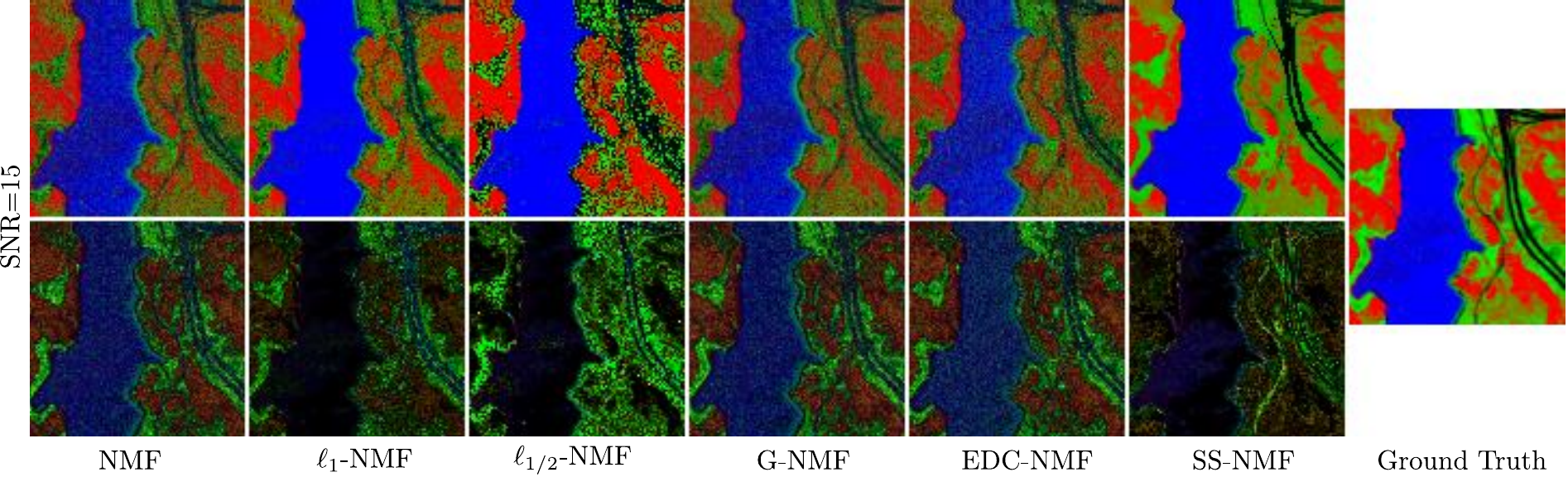}} 
\par\end{centering}

\begin{centering}
\subfloat[\emph{Abundance }maps in gray scale.]{\centering{}\includegraphics[width=0.99\columnwidth]{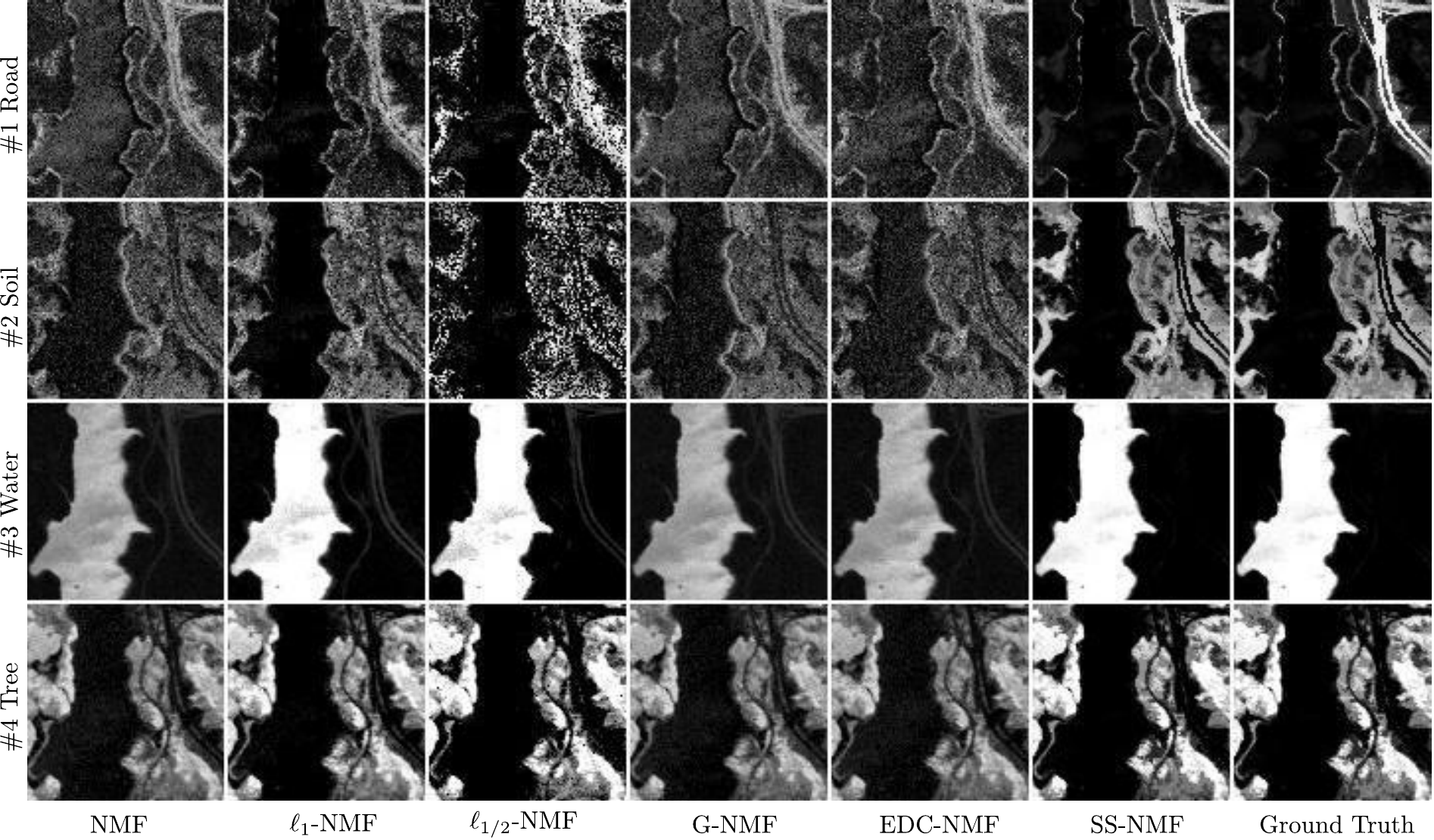}}
\par\end{centering}

\caption{\emph{Abundance }maps under SNR$=15$ dB noise level on the Japer
Ridge data: (a) in pseudo color and (b) in gray scale. There are two
rows in (a). The second row displays the absolute value of the difference
between the estimated results and the ground truth. For each subfigure
in (a), the proportions of Red, Blue, Green and Black ink associated
with each pixel represent the fractional \emph{abundances} of Tree,
Water, Soil and Road in the corresponding pixel. There are four rows
and seven columns in (b). Each row shows the \emph{abundance }maps
of one target. From the $1$-st to the $6$-th column, each column
illustrates the results for one algorithm. The last column shows the
Ground Truths. ``\#1 Road'' in the first row and ``\#2 Soil'' in the
second row in (b) are two difficult targets due to the similarity
spectral curves between them. \label{fig:Abundance_JasperTwoWays}}
\end{figure}

In order to give an intuitive comparison of the HU results, we illustrate
the \emph{abundance} maps on Urban data in Figs.~\ref{fig:Abundance_UrbanAllSNRS},~\ref{fig:Abundance_UrbanTwoWays}
and the \emph{abundance} maps on Japer Ridge data in Figs.~\ref{fig:Abundance_JasperAllSNRS},~\ref{fig:Abundance_JasperTwoWays}.
There are two ways to show the \emph{abundance} maps: in pseudo color
(as shown in Figs.~\ref{fig:Abundance_UrbanAllSNRS},~\ref{fig:Abundance_UrbanTwoWays}a,~\ref{fig:Abundance_JasperAllSNRS},
and~\ref{fig:Abundance_JasperTwoWays}a) and in gray scale (as shown
in Figs.~\ref{fig:Abundance_UrbanTwoWays}b and~\ref{fig:Abundance_JasperTwoWays}b).
Taking the last subfigure in Fig.~\ref{fig:Abundance_UrbanTwoWays}a
for example, we illustrate fractional \emph{abundances} $A_{kn}$
associated with pixel $\mtbldy_{n}$ by plotting the corresponding
pixel using proportions of Red, Blue, Green and Black ink given by
$A_{kn}$ for $k=1,2,3,4$, respectively \cite{Bishop_06_Springer_PRML}.
So, for instance, a pixel for which $A_{1n}=1$ will be colored red,
whereas one for which $A_{1n}=A_{2n}=0.5$ will be colored with equal
proportions of Red and Blue ink and so will appear Purple. The Figs.~\ref{fig:Abundance_UrbanAllSNRS},~\ref{fig:Abundance_JasperAllSNRS},
and~\ref{fig:Abundance_JasperTwoWays}a are achieved in this way.
The gray scale way is straightforward. Taking the subfigures in the
last column in Fig.~\ref{fig:Abundance_UrbanTwoWays}b for example,
we illustrate fractional \emph{abundances} $A_{kn}$ associated with
pixel $\mtbldy_{n}$ by plotting the corresponding pixels in the four
subfigures with different gray scale appearances given by $A_{kn}$
for $k=1,2,3,4$ respectively. So, for instance, a pixel for which
$A_{1n}=1$ will be white in the first subfigure and black in the
other three subfigures. 

Comparing the results in Figs.~\ref{fig:Abundance_UrbanAllSNRS},~\ref{fig:Abundance_UrbanTwoWays},~\ref{fig:Abundance_JasperAllSNRS}
and~\ref{fig:Abundance_JasperTwoWays}, we have
\begin{itemize}
\item SS-NMF achieves the most similar \emph{abundance} maps according to
the Ground Truths. In addition, the colors in our results are regional
smooth and sparsely mixed by Red, Blue, Green and Black ink. These
demonstrate that the structured sparse constraint is effective and
meaningful.
\item \emph{Abundance} maps estimated by the sparse regularized NMF methods
($\ell{}_{\text{1}}$-NMF, $\ell{}_{\text{1/2}}$-NMF) contain more
noise than those of SS-NMF and G-NMF. The reason is that the graph
structure constraint is a kind of smooth constraint, which could urge
the learnt \emph{abundance} maps to be smooth.
\begin{figure}[h]
\begin{centering}
\subfloat[Urban data.]{\begin{centering}
\includegraphics[width=0.33\columnwidth]{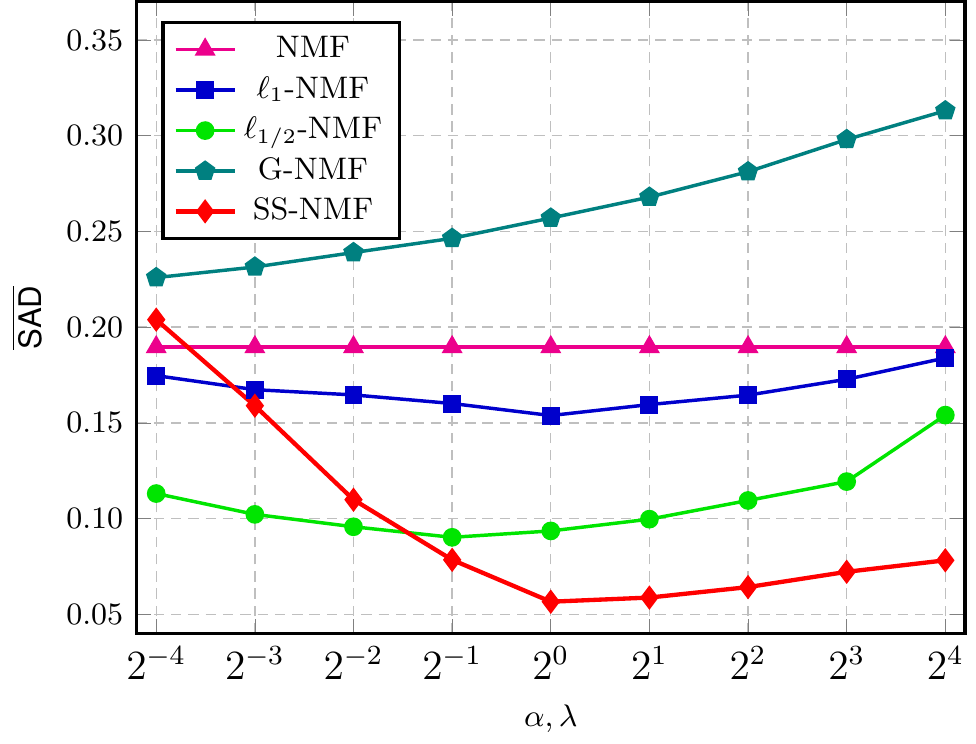}\hspace{0.6cm}\includegraphics[width=0.33\columnwidth]{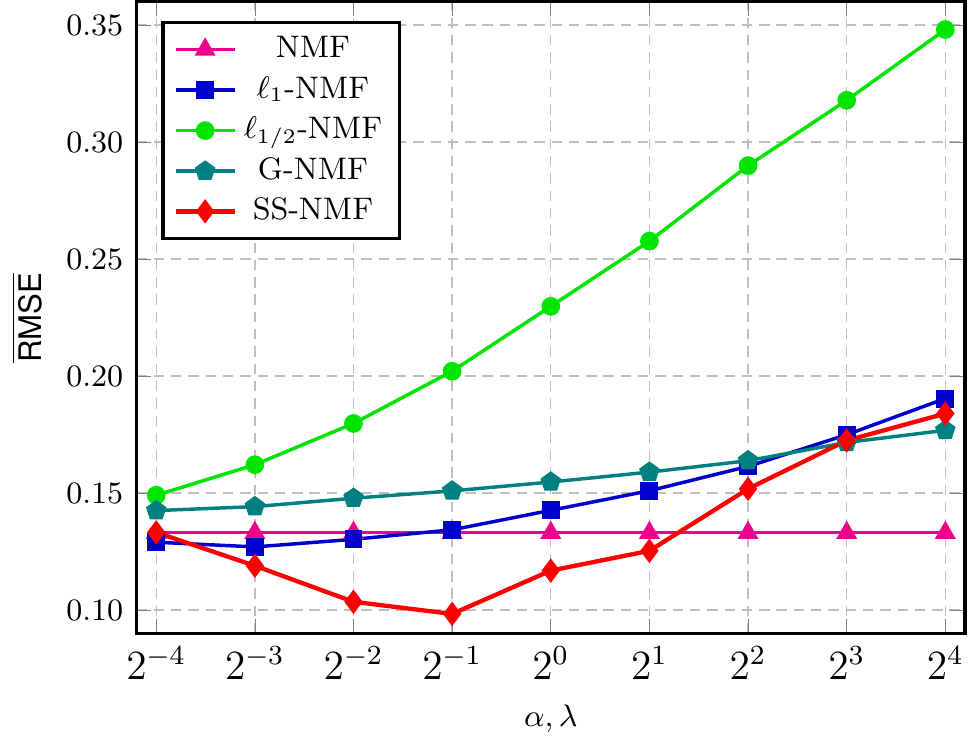}
\par\end{centering}

}
\par\end{centering}

\begin{centering}
\subfloat[Jasper Ridge data.]{\begin{centering}
\includegraphics[width=0.33\columnwidth]{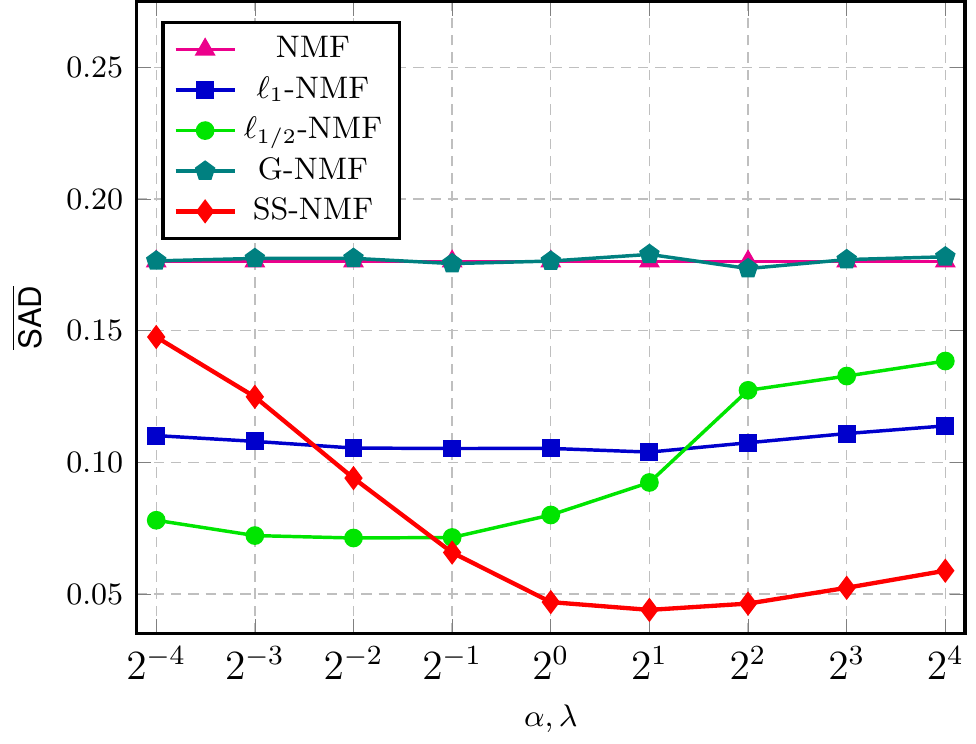}\hspace{0.6cm}\includegraphics[width=0.33\columnwidth]{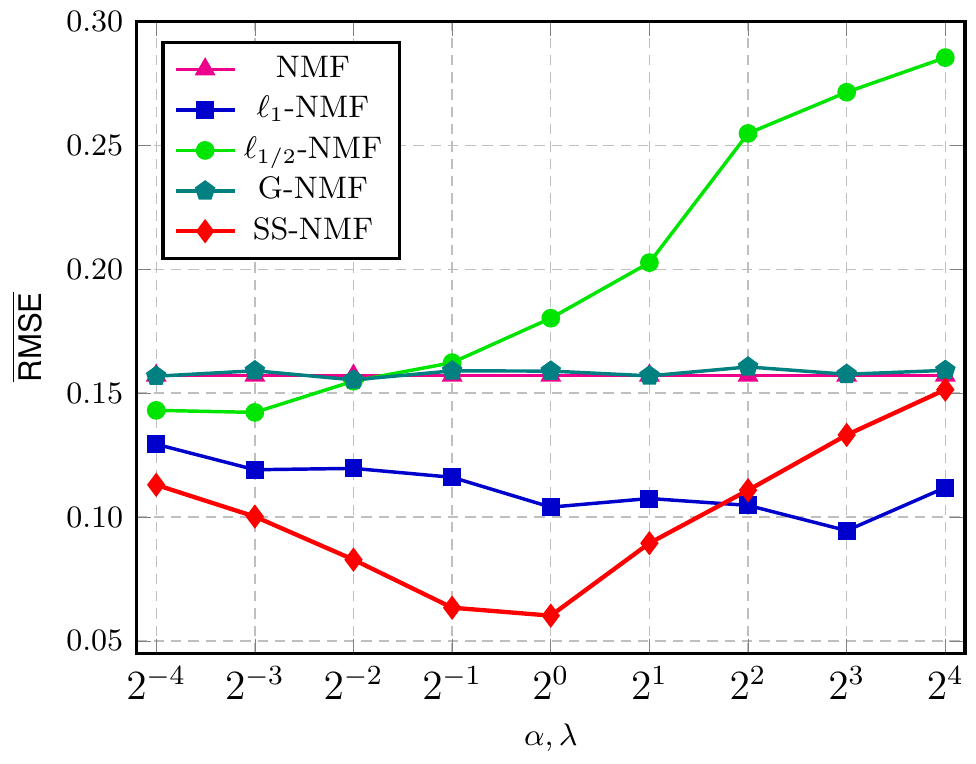}
\par\end{centering}

}
\par\end{centering}

\centering{}\caption{Performance vs. the varying parameters $\lambda\ \text{and}\ \alpha$:
(a) on Urban data, (b) on Jasper Ridge data. The first column shows
the $\overline{\text{SAD}}$s, while the second column displays the
$\overline{\text{RMSE}}$s. The symbol `$2^{4}$' at the bottom-right
of each subfigure indicates $\alpha,\lambda$ equal to $2^{4}$ times
of the values of the best parameter setting respectively.\label{fig:parametersVary}}
\end{figure}

\end{itemize}

\subsection{The Performance versus varying parameters}

In this subsection, we evaluate the impact of varying regularized
parameters upon the performances. Nine experiments are conducted with
respect to nine different regularized parameters on the hyperspectral
images, which are not degraded by manually adding Gaussian noise.
The regularized parameters vary in the following way: $\lambda=$
$2^{-4}\hat{\lambda}$, $2^{-3}\hat{\lambda}$, $2^{-2}\hat{\lambda}$,
$2^{-1}\hat{\lambda}$, $2^{0}\hat{\lambda}$, $2^{1}\hat{\lambda}$,
$2^{2}\hat{\lambda}$, $2^{3}\hat{\lambda}$, $2^{4}\hat{\lambda}$
and $\alpha=2^{-4}\hat{\alpha}$, $2^{-3}\hat{\alpha}$, $2^{-2}\hat{\alpha}$,
$2^{-1}\hat{\alpha}$, $2^{0}\hat{\alpha}$, $2^{1}\hat{\alpha}$,
$2^{2}\hat{\alpha}$, $2^{3}\hat{\alpha}$, $2^{4}\hat{\alpha}$.
Here the values of $\hat{\lambda},\ \hat{\alpha}$ are the ones that
help SS-NMF to get the best results on each data set. Obviously, they
are different with respect to the two data sets. To reduce randomness
of the results, each experiment is repeated ten times and the mean
results are provided. 

Fig.~\ref{fig:parametersVary} shows the graphical performances:
(a) on Urban data, (b) on Jasper Ridge data. For simplicity, the average
performances are provided for each hyperspectral image: $\overline{\text{SAD}}$
on the left side and $\overline{\text{RMSE}}$ on the right side.
Taking the Fig.~\ref{fig:parametersVary}a for example, each value
in the left subfigure is achieved by performing an average calculation
of the SADs of all estimated \emph{endmembers};\emph{ }Each value
in the right subfigure is obtained by averaging the RMSEs of all estimated
\emph{abundance }maps. There is no regularized parameter in NMF, leading
to the graphical results of NMF to be plain. From the results in Fig.~\ref{fig:parametersVary},
we have 
\begin{itemize}
\item As the regularized parameters $\left\{ \lambda,\alpha\right\} $ vary
from $2^{-4}\left\{ \hat{\lambda},\hat{\alpha}\right\} $ to $2^{4}\left\{ \hat{\lambda},\hat{\alpha}\right\} $,
the performances of SS-NMF increase first and then decrease. The proposed
SS-NMF gets good performance in the interval of $\lambda=\left[2^{-2},2^{2}\right]\hat{\lambda}$,
$\alpha=\left[2^{-2},2^{2}\right]\hat{\alpha}$. Especially, SS-NMF
outperforms the other algorithms greatly under suitable parameter
setting $\lambda=\hat{\lambda},\alpha=\hat{\alpha}$. 
\item It seems $\ell_{1}$-NMF, $\ell_{1/2}$-NMF and G-NMF are more robust
to the varying regularized parameters than our SS-NMF method. There
might be two reasons for this phenomenon. First, the results of $\ell_{1}$-NMF,
$\ell_{1/2}$-NMF and G-NMF are much worse than that of our method.
That is, the varying regularized parameters cannot make these algorithms
to get much better or much worse results under different parameter
conditions. Second, although our SS-NMF method could achieve significantly
better results than the other methods, it achieves bad results under
extreme parameter settings. 
\begin{figure}[tbh]
\begin{centering}
\includegraphics[width=0.72\columnwidth]{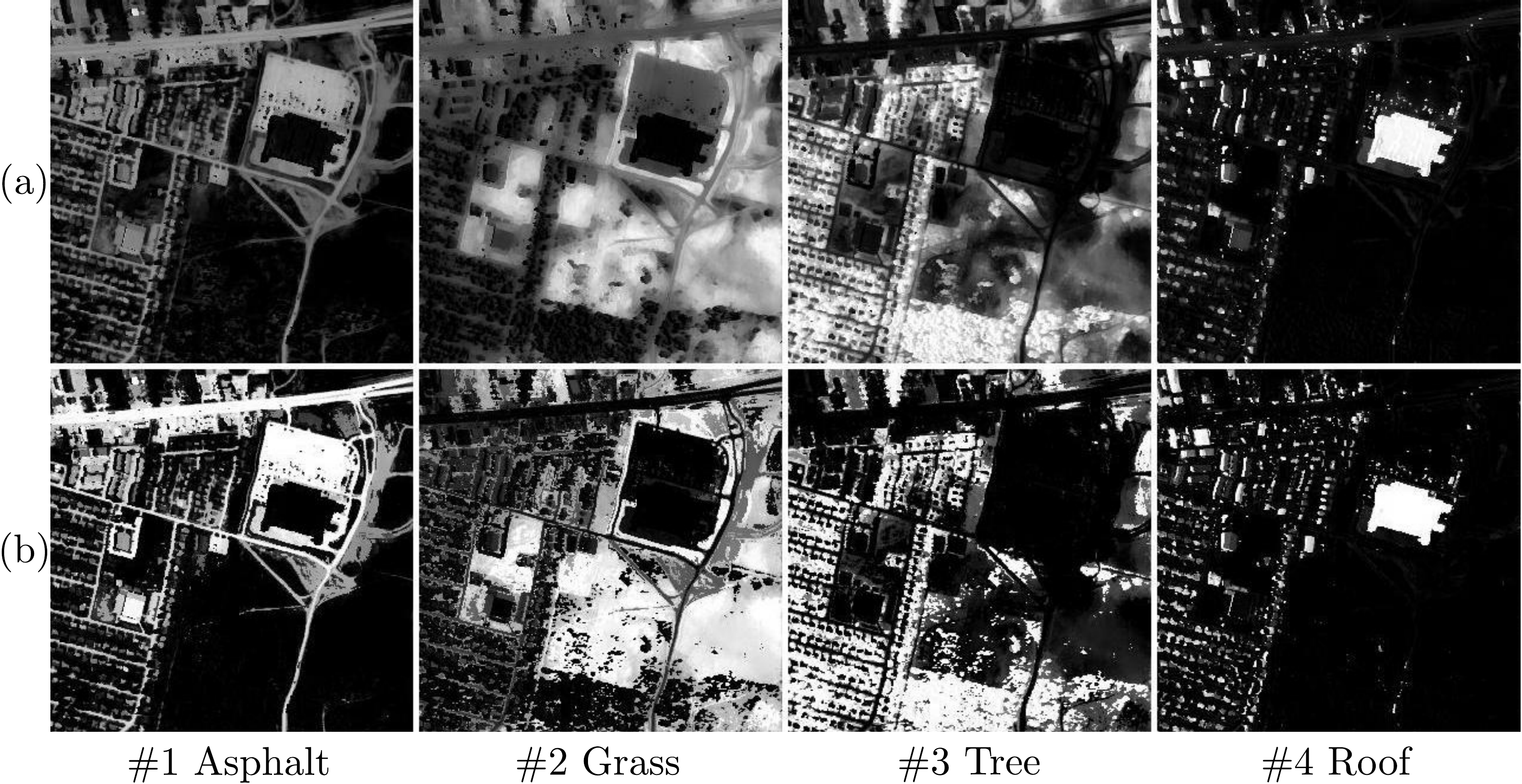}
\par\end{centering}

\caption{The performances of SS-NMF vs. two weighting methods on Urban data.
(a) Only the spatial distance is considered when the local graph structures
are constructed. (b) Both spatial and feature distances are considered
when the local structures are constructed. Each column shows one kind
of \emph{abundance }maps respectively, i.e. \#1 Asphalt, \#2 Grass,
\#3 Tree and \#4 Roof. \label{fig:twoWeighting_urban}}
\end{figure}
 
\begin{figure}[tbh]
\begin{centering}
\includegraphics[width=0.72\columnwidth]{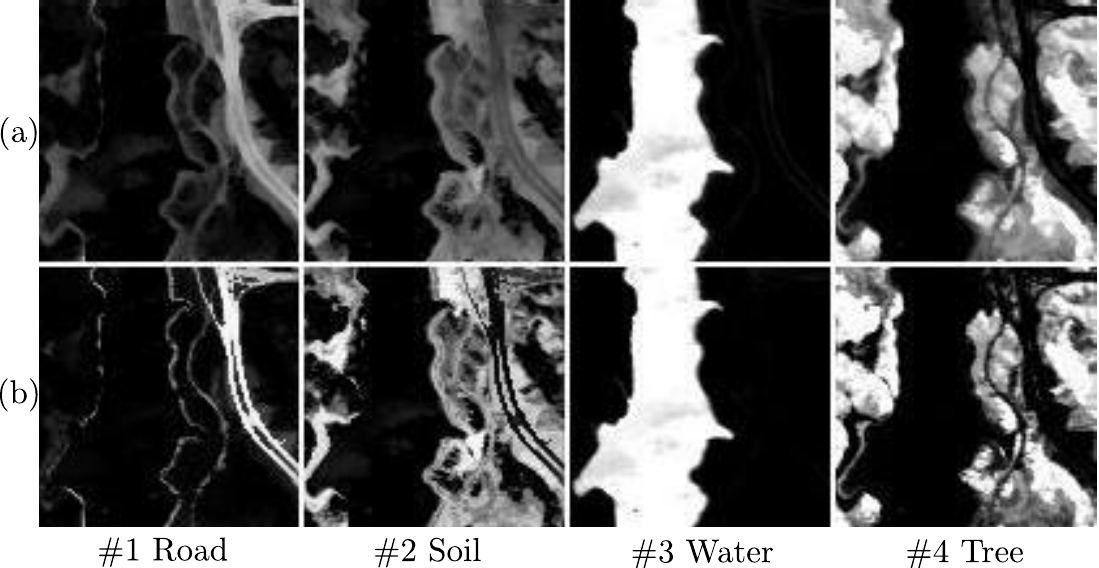}
\par\end{centering}

\caption{The performances of SS-NMF vs. two weighting methods on Jasper Ridge
data. (a) Only the spatial distance is considered when the local graph
structures are constructed. (b) Both spatial and feature distances
are considered when the local structures are constructed. Each column
shows one kind of \emph{abundance }maps respectively, i.e. \#1 Road,
\#2 Soil, \#3 Water and \#4 Tree. \label{fig:twoWeighting_Jasper}}
\end{figure}
 
\end{itemize}

\subsection{Influences of Weighting Methods }

There are many weighting methods. Three common ones are considering
the feature distance only, the spatial distance only and the spatial
and feature distances simultaneously. The second and the third methods
are well suited for hyperspectral images. The reason is that hyperspectral
images enjoy the geometrically intuitive property, that is, the pixels
are located on the grids of a 2-D plane, which means the spatial neighborhood
is inherent for each pixel. In the previous experiments, we employ
the third weighting method, which treats the pixels among the 30\%
biggest SADs in the local windows as nearest neighbors. In this subsection,
the HU performances of the second and the third weighting methods
are thoroughly compared. 

Figs.~\ref{fig:twoWeighting_urban} and~\ref{fig:twoWeighting_Jasper}
show the \emph{abundance }maps obtained by the SS-NMF algorithm with
different weighting methods: (a) considering the spatial distance
only and (b) considering the spatial and feature distances simultaneously.
This experiment is conducted on the hyperspectral images, which are
not degraded by manually adding Gaussian noise. As Figs.~\ref{fig:twoWeighting_urban}a
and~\ref{fig:twoWeighting_Jasper}a show, the results of (a) weighting
method are too smooth and hazy. The reason is that only considering
the spatial distance easily urges a lot of pixels with dissimilar
spectral signatures to be connected on the graph. Accordingly, minimizing
the objective function transfers these graph constraints to the \emph{abundance
}space. Besides, this part of improper constraints would confuse the
optimization algorithm when solving the nonconvex problem, resulting
in bad minima.

\subsection{Convergence Study \label{sub:Convergence-Study}}

The objective function for SS-NMF has been proved non-increasing under
the updating rules~\eqref{eq:updatingRule_M} and~\eqref{eq:updatingRule_A}.
In this subsection, we study the convergence rate and convergence
time for the proposed algorithm. Fig.~\ref{fig:ConvergenceCurves}
illustrates the convergence curves of NMF and SS-NMF on the two hyperspectral
data sets: (a) on Urban data set and (b) on Jasper Ridge data set.
In each subfigure, the $x$-axis shows the number of iterations and
the $y$-axis displays the energies of the objective functions. As
can be seen, the updating rules for both algorithms are efficient,
usually within 50 iterations; SS-NMF converges a bit faster, converging
within 30 iterations.

The convergence time, measured in seconds, for NMF and SS-NMF is summarized
in Table~\ref{tab:convergingTime}. There are three rows. The first
row shows the time of constructing weighted graphs. Since NMF does
not need this part, the time of constructing graphs is zero for this
method, which is denoted by ``--'' in the table. The second row shows
the time of iteration; The total convergence time is summarized in
the last row. Comparing the results in Table~\ref{tab:convergingTime},
we have
\begin{itemize}
\item The SS-NMF method costs less time than the NMF method does during
the iteration process. The reason is that SS-NMF needs less iterations
than NMF does before convergence, as shown in Fig.~\ref{fig:ConvergenceCurves}. 
\item Although NMF costs less convergence time than SS-NMF does on the Urban
data, it fails to continue this superiority on the Jasper Ridge data.
The reason might be that the size of Jasper Ridge data is much smaller
than that of the urban data, approximately $\frac{1}{9}$. It leads
to the ratio of $\frac{t_{\text{construct graph}}}{t_{\text{convergence}}}$
to be smaller on the Jasper Ridge data than that on the Urban data,
i.e., $0.30$ and $0.38$ respectively. 
\begin{figure}[tbh]
\begin{centering}
\subfloat[Urban]{\begin{centering}
\includegraphics[width=0.24\columnwidth]{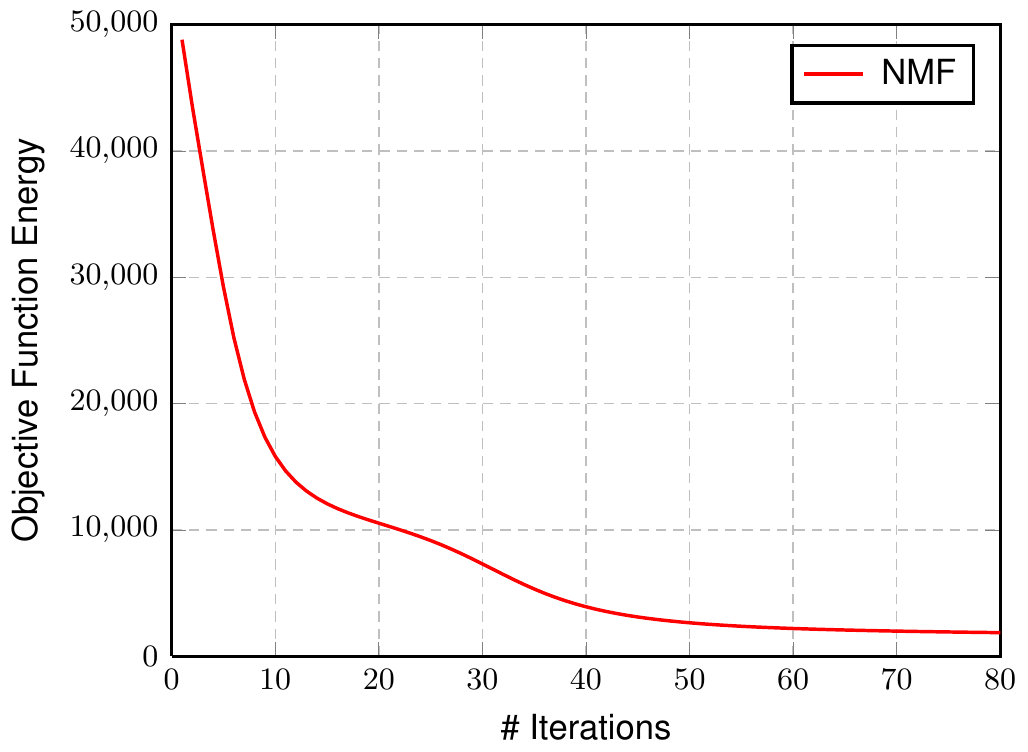}\includegraphics[width=0.24\columnwidth]{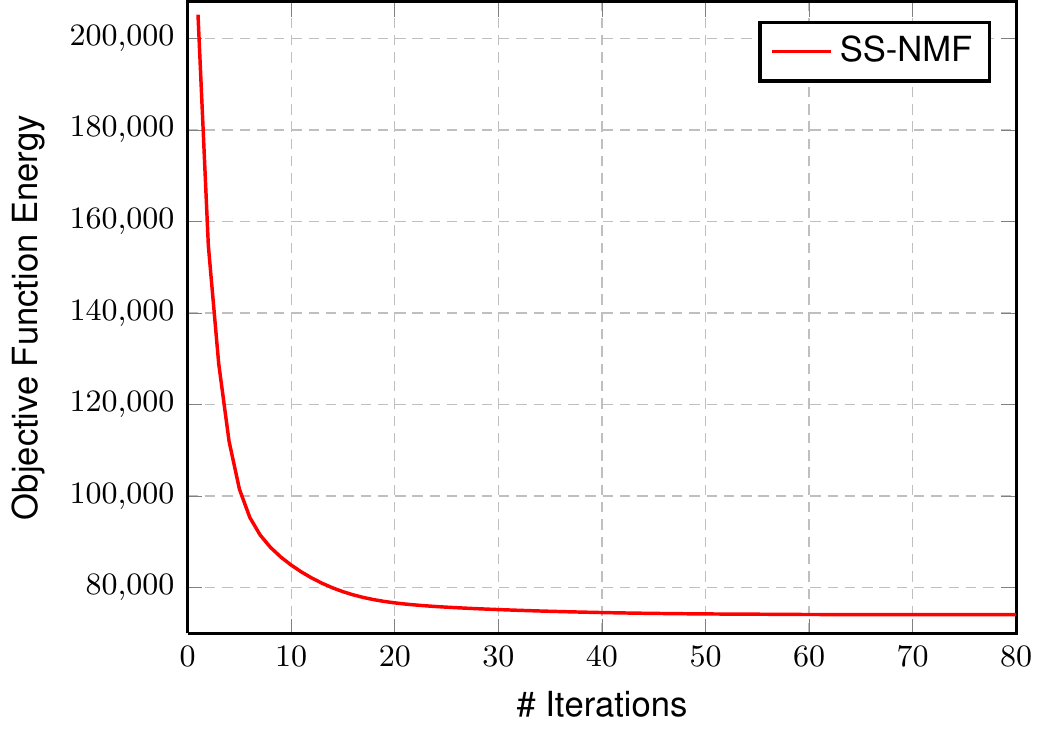}
\par\end{centering}

}\subfloat[Jasper Ridge]{\begin{centering}
\includegraphics[width=0.24\columnwidth]{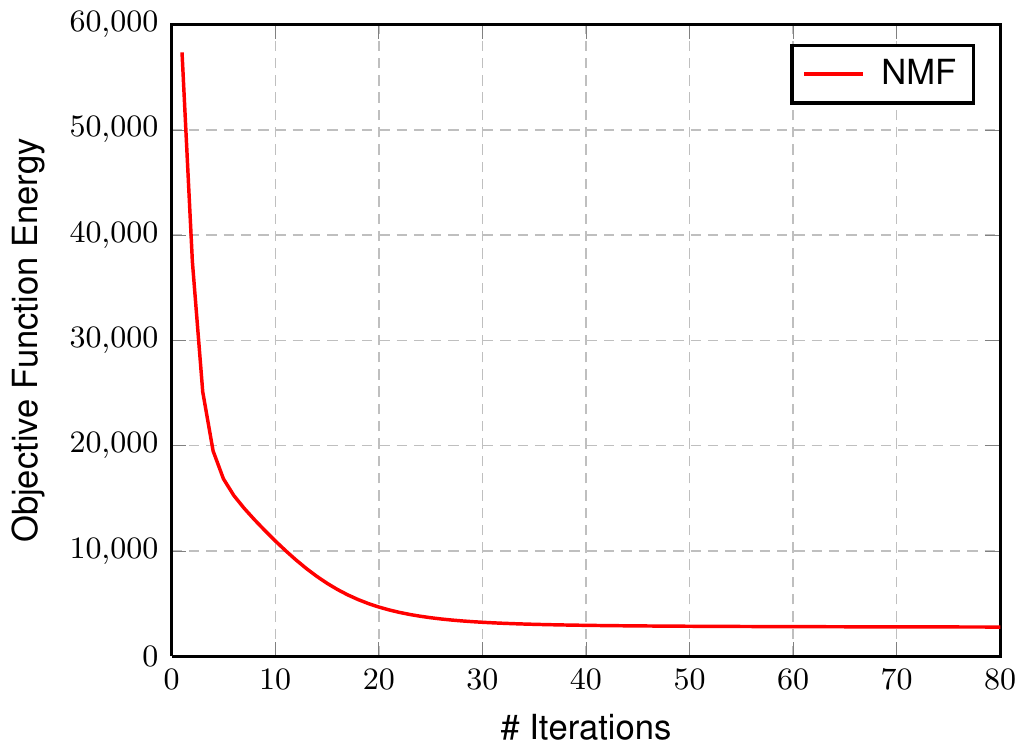}\includegraphics[width=0.24\columnwidth]{31_media_data_important_doneWork_myWorks_isprs_2013_fig_ConverPlot_jasper_ssnmf.pdf}
\par\end{centering}

}
\par\end{centering}

\caption{Convergence curves of NMF and SS-NMF. \label{fig:ConvergenceCurves}}
\end{figure}
 
\begin{table}[tb]
\noindent \centering{}\caption{\label{tab:convergingTime}The time (in second) of constructing graph
\& iteration \& convergence on the two data sets.}
{\small }%
\begin{tabular}{|c||c|c|c|c|}
\hline 
\multirow{2}{*}{{\small Time ($s$)}} &
\multicolumn{2}{c|}{{\small Urban data set}} &
\multicolumn{2}{c|}{{\small Jasper Ridge data set}}\tabularnewline
\cline{2-5} 
 & {\small NMF} &
{\small SS-NMF} &
{\small NMF} &
{\small SS-NMF}\tabularnewline
\hline 
{\small $t_{\text{construct graph}}$} &
{\small --} &
{\small $5.45\pm0.01$} &
{\small --} &
{\small $0.60\pm0.01$}\tabularnewline
\hline 
{\small $t_{\text{iteration}}$} &
{\small $14.03\pm0.12$} &
\textbf{\textcolor{red}{\small $8.83\pm0.11$}} &
{\small $2.19\pm0.02$} &
\textbf{\textcolor{red}{\small $1.39\pm0.13$}}\tabularnewline
\hline 
{\small $t_{\text{convergence}}$} &
\textbf{\textcolor{red}{\small $14.03\pm0.12$}} &
{\small $14.28\pm0.11$} &
{\small $2.19\pm0.02$} &
\textbf{\textcolor{red}{\small $1.99\pm0.14$}}\tabularnewline
\hline 
\end{tabular}
\end{table}
 
\end{itemize}

\section{Conclusion}

Based on the observed properties of hyperspectral data, we propose
an effective method for the HU problem by imposing a Structured Sparse
constraint in the NMF framework, which is called SS-NMF for short.
SS-NMF can effectively overcome the three limitations of NMF for HU
problem. With the structured sparse constraint, SS-NMF not only transfers
the manifold structures inherently embedded in the original data space
to the learned \emph{abundance} space, but also learns an expressive
and accurate set of \emph{endmembers}. Experiments on several hyperspectral
data sets show that our method achieves better results than the state-of-the-art
methods in the sense of both quantitative and qualitative performances.
Besides, our method is relatively robust to different noise levels.
 

{\small \bibliographystyle{IEEEtran}
\bibliography{33_home_zfy_important_doingWork_referenceBib}
}
\end{document}